\documentclass[11pt]{article}

\usepackage[final]{acl}
\usepackage{times}
\usepackage{latexsym}
\usepackage{helvet}  % DO NOT CHANGE THIS
\usepackage{courier}  % DO NOT CHANGE THIS
\usepackage{amsmath}
\usepackage{amsfonts} % for \mathbb
\usepackage{booktabs}
\usepackage{multirow} 
\usepackage{array}
\usepackage{xcolor}
\usepackage{makecell}
\usepackage{natbib}  % DO NOT CHANGE THIS AND DO NOT ADD ANY OPTIONS TO IT
\usepackage{caption} % DO NOT CHANGE THIS AND DO NOT ADD ANY OPTIONS TO IT
\usepackage{float}
\usepackage{multicol}
\usepackage{tcolorbox}

\usepackage[T1]{fontenc}
\usepackage[utf8]{inputenc}

\usepackage{microtype}

\usepackage{inconsolata}

\usepackage{graphicx}

\title{KidsArtBench: Multi-Dimensional Children's Art \\
Evaluation with Attribute-Aware MLLMs} %A Pedagogically-Grounded Benchmark for Children's Art Evaluation with Attribute-Aware MLLMs 

\author{
\textbf{Mingrui Ye\textsuperscript{1}},
\textbf{Chanjin Zheng\textsuperscript{2}},
\textbf{Zengyi Yu\textsuperscript{2}},
\textbf{Chenyu Xiang\textsuperscript{3}}, \\
\textbf{Zhixue Zhao\textsuperscript{3}},
\textbf{Zheng Yuan\textsuperscript{3}},
\textbf{Helen Yannakoudakis\textsuperscript{1}} \\
\textsuperscript{1}King's College London \quad
\textsuperscript{2}East China Normal University \quad
\textsuperscript{3}University of Sheffield  \\
\texttt{\{mingrui.ye, helen.yannakoudakis\}@kcl.ac.uk}, \texttt{chjzheng@dep.ecnu.edu.cn}, \\
\texttt{51284118014@stu.ecnu.edu.cn}, \texttt{\{cxiang10, zhixue.zhao, zheng.yuan1\}@sheffield.ac.uk}
}

\begin{document}
\maketitle

\begin{abstract}
Multimodal Large Language Models (MLLMs) show remarkable progress across many visual–language tasks; however, their capacity to evaluate artistic expression remains limited: aesthetic concepts are inherently abstract and open-ended, and multimodal artwork annotations are scarce. We introduce KidsArtBench, a new benchmark of over 1k children's artworks (ages 5-15) annotated by 12 expert educators across 9 rubric-aligned dimensions, together with expert comments for feedback. Unlike prior aesthetic datasets that provide single scalar scores on adult imagery, KidsArtBench targets children's artwork and pairs multi-dimensional annotations with comment supervision to enable both ordinal assessment and formative feedback. Building on this resource, we propose an attribute-specific multi-LoRA approach -- where each attribute corresponds to a distinct evaluation dimension (e.g., Realism, Imagination) in the scoring rubric -- with Regression-Aware Fine-Tuning (RAFT) to align predictions with ordinal scales. On Qwen2.5-VL-7B, our method increases correlation from 0.468 to 0.653, with the largest gains on perceptual dimensions and narrowed gaps on higher-order attributes. These results show that educator-aligned supervision and attribute-aware training yield pedagogically meaningful evaluations and establish a rigorous testbed for sustained progress in educational AI. We release data and code with ethics documentation.\footnote{\url{https://github.com/bigrayss/KidsArtBench}}

%%We further evaluate the benchmark using open-source MLLMs (e.g., Qwen2.5-VL family, Gemma3 family, Mimo-VL, and Kimi-a3b). Building on this, we introduce the ArtMentor base model for MLLMs that integrates attribute-specific Low‐Rank Adaptation (LoRA) and Regression-Aware Fine-Tuning (RAFT). By leveraging our curated dataset, MLLMs can achieve significantly improved performance in student art assessment, highlighting their potential to provide pedagogically meaningful evaluations of student artwork and contributing both methodological advances and valuable resources for educational AI.

\end{abstract}

\section{Introduction}

%%Recently, Multimodal Large Language Models (MLLMs) have significantly transformed various domains, with education being one of the most promising fields~\cite{11016818, qian2025pedagogical}. MLLMs, thanks to their capability to understand and generate information across multiple modalities, offer unprecedented opportunities to improve educational processes by improving assessment and feedback mechanisms and by supporting and empowering learning experiences~\cite{xing2024survey, lee2024llava}. Among all, art evaluation and creativity assessment play significant role in education which are essential for fostering innovation, self-expression, and skill development~\cite{denac2014significance, robson2012observing}. By providing structured feedback, these assessments help students refine their artistic techniques while encouraging originality and deeper creative thinking~\cite{seo2022toward, zhao2024revolutionizing}.

Multimodal Large Language Models (MLLMs) have shown impressive capabilities across visual-language tasks such as captioning, reasoning, and instruction following~\cite{achiam2023gpt,team2024gemini,biswas2024robustness}. In education, MLLMs offer the potential to transform assessment and feedback workflows across modalities~\cite{repec:gam:jftint:v:16:y:2024:i:12:p:467-:d:1543466,lee2024llava}, enhancing accessibility, scalability, and personalization.

Among educational tasks, evaluating student artwork remains especially challenging. Artistic expression is inherently abstract and subjective~\cite{zhu2021learning}. This issue is further compounded by the scarcity of human-annotated multimodal artwork data, limiting MLLMs' ability to accurately perceive and assess artworks~\cite{huang2024aesbench}. Furthermore, traditional evaluation methods often rely on labor-intensive, inconsistent human judgment~\cite{sali2014analysis,meyer2024algorithmic}. While recent research has explored deep learning approaches to aesthetics~\cite{jiang2024aacp,she2021hierarchical}, these models tend to collapse creativity into a single scalar score, lack fine-grained interpretability, and alignment with educational or pedagogical goals~\cite{huang2024aesbench}.

\begin{figure}[t]
  \centering
  \includegraphics[scale=0.25]{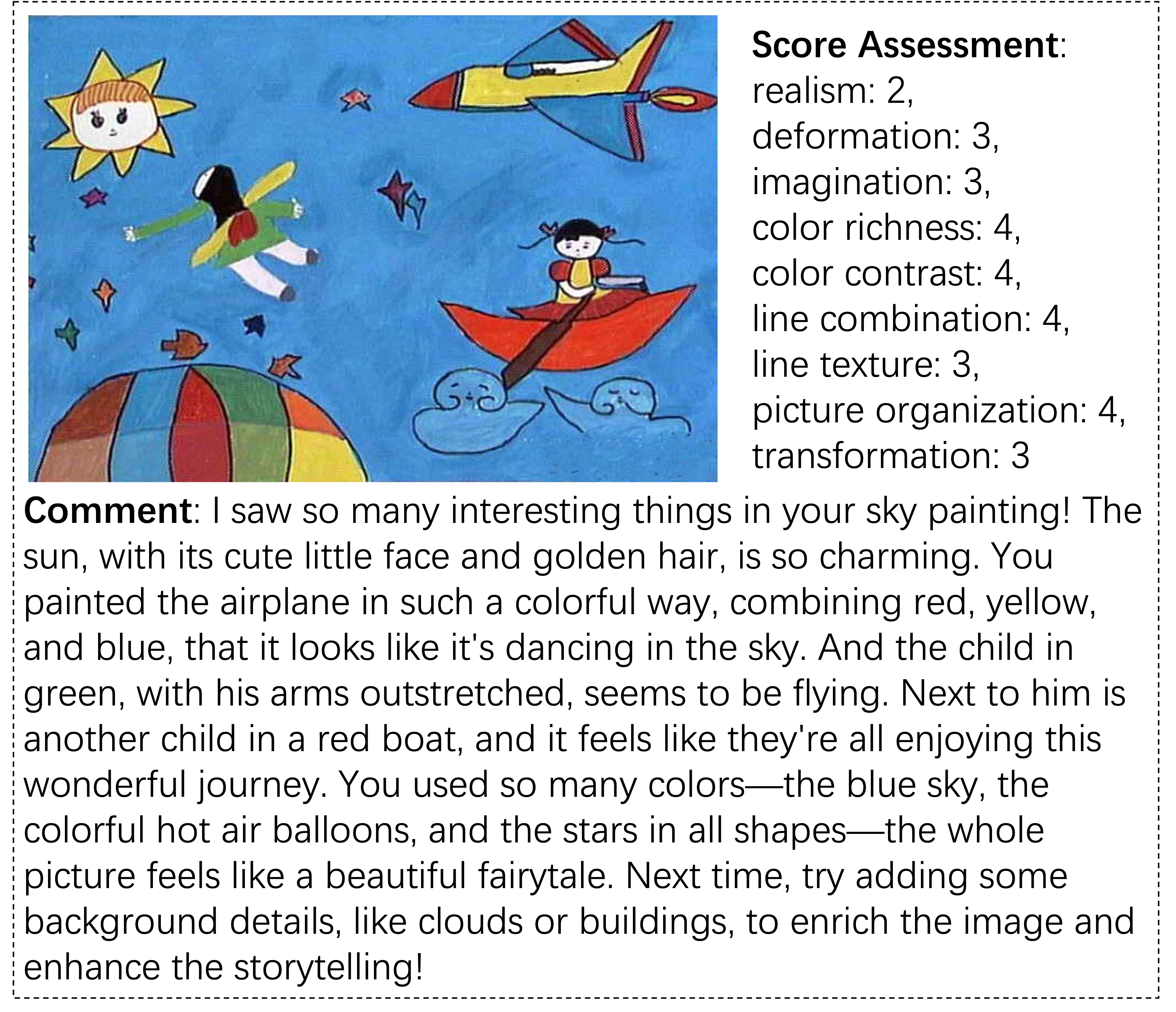}
  \caption{Sample from the KidsArtBench dataset. Each artwork includes 9 rubric-based scores and (in a subset) educator comments.}
  \label{fig:fig1}
\end{figure}

\begin{figure*}[htbp]
  \centering
  \includegraphics[scale=0.2]{fig2.png}
  \caption{Overview of our framework. Left: prompting MLLMs with rubric-aligned instructions. Right: multi-LoRA architecture with RAFT/RAIL for score prediction.}
  \label{fig:fig2}
\end{figure*}

However, structured visual evaluation -- particularly in art education -- plays a critical role in fostering self-expression, technical skill, and creative development~\cite{denac2014significance,robson2012observing,seo2022toward,zhao2024revolutionizing}. Effective feedback not only assesses surface-level features (e.g., color use, line structure) but also supports higher-order reflection on originality, transformation, and composition. This calls for models capable of nuanced, dimension-specific, educator-aligned evaluation.

%%Traditional methods~\cite{sali2014analysis} for evaluating artworks rely heavily on manual judgment, which can be subjective and inconsistent. Similarly, Human–Computer Interaction (HCI) based approaches~\cite{meyer2024algorithmic} also share these limitations, but offer the advantage of enhancing art exploration by linking visual perception with interactive experience. Researchers further have explored deep learning methods~\cite{jiang2024aacp, she2021hierarchical} to assess the aesthetic quality of paintings which could be more efficient and objective, while their practical utility is severely constrained by the lack of transferability and generalizability. In contrast, MLLMs can fuse visual and linguistic knowledge enable a comprehensive, human-like analysis of artworks, assessing aspects from low-level aesthetics scores to higher-order elements like comments and thematic content. However, the abstract nature of artwork aesthetics perception~\cite{zhu2021learning} presents a significant challenge for current MLLMs. This issue is further compounded by the scarcity of human-annotated multimodal artwork data, limiting MLLMs' ability to accurately perceive and assess students artworks~\cite{huang2024aesbench}.

\noindent \textbf{Our contributions.}  
We present KidsArtBench, a new benchmark for multi-dimensional evaluation of children's artwork (ages 5–15), comprising 1,046 student submissions scored by 12 experienced art educators across 9 standardized rubric dimensions (e.g., Realism, Color Richness, Imagination). We refer to these dimensions as \textit{attributes} throughout the paper, aligning with the terminology of multi-attribute modeling in vision-language learning. In contrast to prior datasets that provide only scalar ratings or aesthetics labels (e.g., AVA~\cite{murray2012ava}, AADB~\cite{kong2016photo}, ACPP~\cite{jiang2024aacp}), KidsArtBench captures a richer educational perspective, with both quantitative scores and qualitative feedback comments. Our work is the first to curate a dataset for art assessment that provides comprehensive point-wise evaluations across multiple dimensions together with pedagogically oriented feedback, as illustrated in Figure~\ref{fig:fig1}.

To evaluate and improve MLLMs on this task, we first prompt open-source MLLMs using rubric-aligned instructions and observe consistent underperformance, especially on abstract or compositional dimensions (e.g., Line Combination, Imagination). We then propose an attribute-aware fine-tuning framework that incorporates: a multi-branch LoRA architecture~\cite{wang2023multilora}, where each module specializes in a distinct rubric dimension; Regression-Aware Fine-Tuning (RAFT)~\cite{lukasik2025better}; and Regression-Aware Inference (RAIL)~\cite{lukasik-etal-2024-regression}, aligning predictions with ordinal scales by performing logits-level learning and applying minimum Bayes risk decoding objective. Figure~\ref{fig:fig2} illustrates our full pipeline, from prompt-based rubric alignment to attribute-specific adapters and score decoding.

On Qwen2.5-VL-7B, our approach improves average correlation from 0.468 (prompted baseline) to 0.653, with substantial gains on both perceptual dimensions (e.g., Realism, Color Richness) and abstract attributes (e.g., Transformation, Picture Organization). Nonetheless, certain dimensions (e.g., Line Texture) remain challenging, underscoring the complexity of structured artistic evaluation -- and highlighting the value of KidsArtBench as a challenging testbed for advancing fine-grained visual understanding in MLLMs.

In summary, our contributions are fourfold: i) We introduce KidsArtBench, the first public benchmark for multi-dimensional evaluation of children's artwork with expert-annotated rubric scores and comments; ii) We propose an attribute-aware MLLM fine-tuning method using multi-LoRA with RAFT, and RAIL to align predictions with ordinal supervision; iii) We show that our method substantially improves MLLM performance on aesthetic assessment tasks, in particular in perceptual dimensions such as Realism and Transformation; iv) We release all data, code, and documentation to support further research in educationally grounded multimodal AI.

%\begin{itemize}
%\item ArtMentor dataset. To our knowledge, ArtMentor dataset is the first to provide comprehensive \textbf{multi-dimensional point-wise metrics} alongside \textbf{pedagogically oriented feedback}, establishing an educationally meaningful benchmark for art assessment.
%\item ArtMentor base model. Our model, built upon Qwen2.5-VL-7B and fine-tuned on our benchmark with the attribute-specific multi-LoRA architecture integrated with RAFT, achieves significant improvements especially for perceptual dimensions like Realism and Transformation.
% Moreover, results from comment-annotated subsets suggest that data augmentation and incorporating comments as auxiliary signals provide additional gains.
%\item Open-source resources. We release both the ArtMentor dataset and the proposed base model to the community. We hope these resources offer a foundation for developing MLLMs with comprehensive aesthetic capabilities, supporting both methodological innovation and educational applications.
%\end{itemize}

\section{Related Work}

%\subsection{Multimodal Large Language Models (MLLMs)}

%%Recent commercial MLLMs such as GPT-series~\cite{achiam2023gpt}, Gemini-series~\cite{team2024gemini}, and Claude-series~\cite{biswas2024robustness} have demonstrated remarkable capabilities in various visual-language tasks, including image recognition, visual question answering, and cross-modal retrieval~\cite{zhang2024mm}. As for the open-sourced MLLMs, Qwen2.5-VL~\cite{bai2025qwen2} advances Vision Language Models (VLMs) through native spatiotemporal processing, efficient windowed ViT, and practical agentic functionalities. Gemma3~\cite{team2025gemma}, which employs optimized local-to-global attention for efficient long-context processing, achieves superior performance through distillation and a novel post-training approach that enables smaller models to match the performance of their larger counterparts. Kimi-VL~\cite{team2025kimi} introduces an efficient MoE-based vision-language model with a 2.8B-parameter activated decoder, featuring native-resolution MoonViT encoding and context processing. Our work builds upon these open-source MLLMs to explore artistic assessment for education.
\paragraph{Multimodal Large Language Models (MLLMs)} Recent commercial MLLMs such as GPT-series, Gemini, and Claude~\cite{achiam2023gpt,team2024gemini,biswas2024robustness}, as well as open-source models like Qwen2.5-VL~\cite{bai2025qwen2}, Qwen3-VL~\cite{qwen3technicalreport}, Gemma3~\cite{team2025gemma}, Kimi-VL~\cite{team2025kimi}, and Mimo-VL ~\cite{coreteam2025mimovltechnicalreport} have achieved strong performance across visual-language tasks including captioning, visual question answering, and cross-modal retrieval~\cite{zhang-etal-2024-mm}. Qwen2.5-VL, for example, incorporates native spatiotemporal processing and efficient windowed ViTs, while Gemma3 employs local-to-global attention for long-context reasoning. 
%%Despite remarkable capabilities, MLLMs still confront several limitations for artistic assessment. Existing work~\cite{li2025benchmark} demonstrates persistent challenges in processing abstract concepts and performing robust causal reasoning. They exhibit notable deficiencies in fine-grained visual tasks, particularly in distinguishing artistic details or domain-specific visual patterns. Furthermore, spatial cognition and relational reasoning capabilities remain underdeveloped in existing architectures~\cite{anis2025limitations} especially for substream tasks. These technical constraints are compounded by the substantial computational requirements that hinder practical applications.
Despite these advances, MLLMs still struggle with fine-grained, multi-attribute evaluation tasks, especially in domains involving abstract or perceptual qualities~\cite{li2025benchmark,anis2025limitations}. They exhibit limited spatial reasoning, inconsistent attention to visual detail, and poor alignment with structured rubrics -- posing major obstacles for use in art assessment and educational settings.

\paragraph{Aesthetic Evaluation and Understanding} Aesthetic evaluation models such as AesCLIP~\cite{sheng2023aesclip} and AesExpert~\cite{huang2024aesexpert} leverage CLIP-like architectures or LLAVA-style prompting for zero-shot scoring or aesthetic question answering. While effective in general image aesthetics, these systems are optimized for scalar preference prediction or aesthetic Q\&A rather than pedagogically grounded, dimension-specific feedback. Moreover, their training data often reflect domain biases (e.g., adult photography or generative art), limiting their applicability to children's work. SemArt~\cite{garcia2018read} and ACPP~\cite{jiang2024aacp} begin to address semantic and child-centered art understanding, respectively. However, they do not provide rubric-aligned, multi-dimensional annotations or comment-level feedback. Existing methods typically collapse artistic merit into a single score, limiting interpretability and educational relevance.

\paragraph{MLLMs in Educational Assessment} LLM-based tools have begun to support automated assessment in text-based education~\cite{bewersdorff2023assessing,latif2024fine}, but multimodal assessment remains underexplored. The ArtMentor framework~\cite{zheng2025artmentor} represents an important step: it uses GPT-4o to generate formative feedback for children's artwork using teacher-in-the-loop evaluation. However, its reliance on proprietary models has raised concerns around transparency, cost, and replicability~\cite{yan2024practical}. In contrast, our work uses open-source MLLMs and contributes both a dataset and fine-tuning framework for rubric-based, pedagogically meaningful visual assessment.

%%Despite constructions of aesthetics datasets (e.g., AVA~\cite{murray2012ava}, AADB~\cite{kong2016photo}, PCCD~\cite{chang2017aesthetic}) and specialized art painting benchmarks (e.g., Art500k~\cite{mao2017deepart}, OmniArt~\cite{strezoski2017omniart}), existing resources are fundamentally inadequate for assessing student artwork in educational contexts. Additionally, most of these datasets are relied on voting systems for final scoring, which inherently lacks explanatory depth regarding the specific aesthetic qualities being assessed. SemArt~\cite{garcia2018read} advances semantic art analysis by incorporating both structured titles and unstructured textual critiques for each artwork, offering a multiple perspective on art interpretation. ACPP~\cite{jiang2024aacp} featuring a comprehensive aesthetics assessment of children's paintings datasets with 1.2K expert-annotated images incorporating eight specialized attributes. In summary, existing aesthetics datasets are critically limited by their lack of analytical breadth, the pervasive presence of generative bias, and their poor alignment with educational needs. Therefore, establishing a comprehensive and context-oriented dataset is crucial for advancing both MLLMs application and educational aesthetic assessment.
\paragraph{Aesthetic and Educational Art Datasets} Many aesthetics datasets exist; e.g., AVA~\cite{murray2012ava}, AADB~\cite{kong2016photo}, PCCD~\cite{chang2017aesthetic}, OmniArt~\cite{10.1145/3273022}, and Art500k~\cite{mao2017deepart}. However, most focus on adult art or photography and rely on aggregate preference ratings (e.g., votes or likes), without explicit rubrics. ACPP~\cite{jiang2024aacp} is one of the few datasets focused on children's art, containing scalar scores across eight attributes. However, it lacks rubric calibration, feedback comments, and modeling benchmarks. Our proposed dataset, uniquely provides expert-annotated, nine-dimensional scores with a rubric-aligned evaluation protocol and a modeling suite designed for both assessment and feedback generation in educational settings. To our knowledge, KidsArtBench is the first dataset to combine expert rubric supervision, multi-dimensional annotation, and open benchmarking protocols for MLLMs in educational art evaluation.

\section{KidsArtBench Dataset}

\begin{table*}[htbp]
\centering
{\scriptsize
\begin{tabular}{@{}lp{6cm}p{5cm}@{}}
\toprule
\textbf{Category} & \textbf{Dimension} & \textbf{Evaluation Criteria} \\
\midrule
\multirow{3}{*}{Formative Creativity} 
 & Realism \cite{biswas2021realism} & Accuracy in depicting subjects and objects \\
 & Deformation \cite{sfarra2014discovering} & Creative reinterpretation of reality \\
 & Imagination \cite{searle2016capturing} & Novelty and originality of concepts \\[0.1cm]

\multirow{2}{*}{Color Expressiveness} 
 & Color Richness \cite{lu2015discovering,pylypchuk2021developing} & Diversity and harmony of color palette \\
 & Color Contrast \cite{zhang2021comprehensive} & Visual impact through hue interactions \\[0.1cm]

\multirow{2}{*}{Line Work Richness} 
 & Line Combination \cite{locher1999empirical} & Structural arrangement of strokes \\
 & Line Texture \cite{ding2020image} & Expressive tactile quality of linework \\[0.1cm]

\multirow{2}{*}{Conceptual Thinking} 
 & Picture Organization \cite{locher1999empirical} & Balanced composition and spatial logic \\
 & Transformation \cite{du2020research} & Effective rendering of abstract concepts \\
\bottomrule
\end{tabular}
}
\caption{KidsArtBench rubric dimensions. Full rubric text in Appendix~\ref{sec:appendix rubrics}.}
\label{tab:table1}
\end{table*}

We collected 1,046 original artworks from students aged 5–15 through an online submission platform designed to encourage authentic creative expression. All data collection followed an approved IRB protocol with parental consent and child assent (see Ethics Statement and Bias section). The dataset primarily consists of artworks collected from children across multiple primary and middle schools in Eastern China. Educator feedback comments were originally written in Chinese; we translated them into English using Google Translate and then manually verified and corrected them with the help of proficient English speakers who are native Chinese speakers. 
We release both the original Chinese comments and their English translations to support a broad range of research communities.

Each artwork was independently scored by at least two trained art educators across nine standardized rubric dimensions (Table~\ref{tab:table1}) on a scale from 1 to 5. A senior expert then adjudicated discrepancies, consulting original raters when necessary to assign a final score for each dimension. This multi-stage process ensured high-quality, educator-aligned annotations. Our dataset also includes expert-written formative comments, providing qualitative guidance aligned with the rubric (Figure~\ref{fig:fig1}).

\begin{figure*}[htbp]
  \centering
  \includegraphics[scale=0.74]{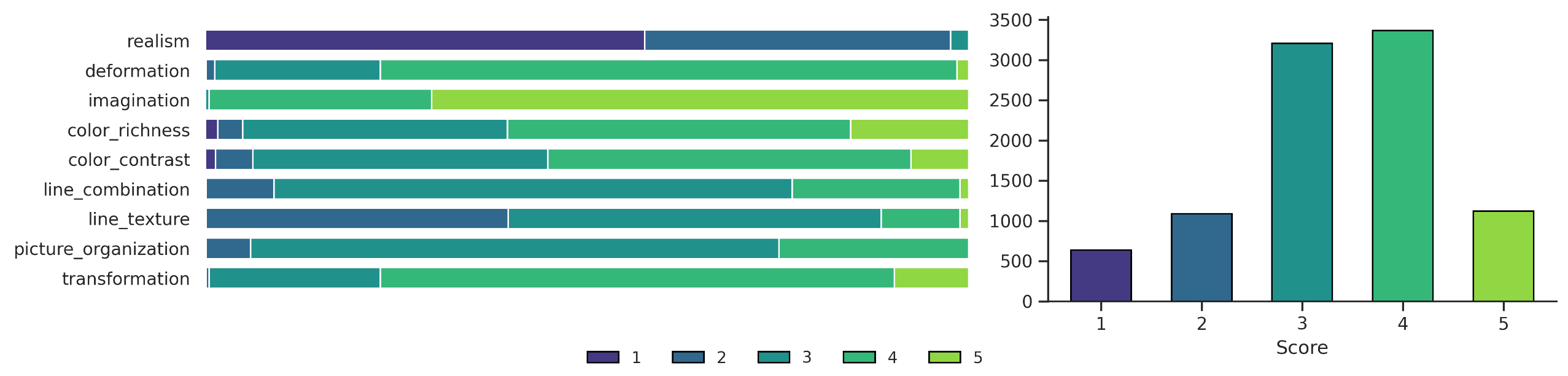}
  \caption{Score distribution across nine dimensions (left) and overall dataset (right) in KidsArtBench.}
  \label{fig:distribution}
\end{figure*}

The dataset is split into 80\% train, 10\% validation, and 10\% test to support model development. Figure~\ref{fig:distribution} shows the score distributions across all dimensions: most scores cluster around 3-4, reflecting realistic assessment tendencies in educational contexts (in our dataset, most student artworks typically demonstrate intermediate to above-average performance), with relatively few extreme values. The rubric captures nuanced variation in artistic quality and provides a representative basis for training and evaluating MLLMs in art education.

\section{Methodology}

Given an artwork image (and optional comment text), the goal is to predict nine ordinal scores \(y_m \in \{1,\dots,5\}\) for each rubric dimension \(m \in \mathcal{D}\) where $ \mathcal{D} = \{\mathrm{realism}, ..., \mathrm{transformation}\}$. This enables dimension-specific assessment rather than a single scalar aesthetic score.

\subsection{Attribute-Aware Prompting with MLLMs}
We explore open-source MLLMs for art evaluation using rubric-guided prompting. Each artwork is assessed using a structured prompt template shown in Figure~\ref{fig:fig2} (left). Full rubric definitions and prompt examples are provided in Appendix~\ref{sec:appendix rubrics}-~\ref{sec:appendix_prompt}. Our attribute-aware prompting strategy enables dimension-specific queries, encouraging the model to reason about aspects such as Realism, or Color Richness. However, our results show that existing MLLMs perform inconsistently on KidsArtBench, especially on abstract and compositional dimensions (e.g., Imagination, Line Combination), motivating the need for fine-tuning with targeted supervision.

\subsection{Attribute-Specific MultiLoRA}
\label{sec:method}

We propose an attribute-aware MLLM fine-tuning framework, in which each evaluation attribute -- corresponding to a rubric dimension such as \textit{Realism} or \textit{Imagination} -- is modeled by a dedicated LoRA adapter. This modular design decomposes the multi-dimensional assessment task into independent, specialized branches, mitigating inter-dimensional interference and enabling fine-grained learning across all criteria~\cite{wang2023multilora,hu2022lora}.

Let $x$ denote the input, which includes the student artwork and optional textual description, and let $\mathbf{z} = f_{0}(x) \in \mathbb{R}^{d}$ be the corresponding embedding from a frozen backbone encoder $f_0$. We define the set of evaluation dimensions as $\mathcal{D}.$  
For each dimension $m \in \mathcal{D}$, we attach a LoRA adapter to generate a scalar prediction $\hat{y}_m$:
\begin{equation}
  \hat{y}_m = \left( \mathbf{W}_0 + \mathbf{B}_m \mathbf{A}_m \right) \mathbf{z} + b_m,
\end{equation}
\noindent where $\mathbf{W}_0 \in \mathbb{R}^{d \times d}$ is the frozen, shared projection matrix; $\mathbf{A}_m \in \mathbb{R}^{r \times d}$ and $\mathbf{B}_m \in \mathbb{R}^{d \times r}$ are trainable low-rank update matrices (LoRA rank $r \ll d$); $b_m \in \mathbb{R}$ is a dimension-specific learnable bias.

This formulation allows each adapter to specialize in one rubric criterion while leveraging the shared visual-linguistic representation $\mathbf{z}$. Importantly, the architecture supports flexible composition: any subset of dimensions $\mathcal{S} \subseteq \mathcal{D}$ can be selectively activated for context-specific evaluation or targeted feedback.

%To decompose nine-dimensional students' artwork assessment, we propose an attribute-specific regression method by employing a dedicated LoRA adapter for each dimension~\cite{wang2023multilora, hu2022lora}. Let $x$ be the input, which includes the student artwork and the description texts, and $\mathbf{z} = f_{0}(x) \in \mathbb{R}^{d}$ be its embedding from a frozen backbone $f_{0}$. We denote the set of dimensions as $ \mathcal{D} = \{\mathrm{realism}, ..., \mathrm{transformation}\}. $ For each dimension $m \in \mathcal{D}$, we give the LoRA adapter scalar prediction results \(\hat{y}_{m}\) as
%\begin{equation}
%  \hat{y}_{m}
%  \;=\;
%  \bigl(\mathbf{W}_{0} + \mathbf{B}_{m}\,\mathbf{A}_{m}\bigr)\,\mathbf{z}
%  \;+\;
%  b_{m},
%\end{equation}
%where \(\mathbf{W}_{0}\in\mathbb{R}^{d\times d}\) denotes the fixed pretrained projection weights shared across all dimensions. \(\mathbf{A}_{m}\in\mathbb{R}^{r\times d}\) and \(\mathbf{B}_{m}\in\mathbb{R}^{d\times r}\) are the trainable low‑rank update matrices with the LoRA rank \(r\ll d\), \(b_{m}\in\mathbb{R}\) is the learnable bias. 

\subsection{Regression-Aware Fine-Tuning (RAFT) and Inference (RAIL)}

To produce well-calibrated ordinal scores, we adopt the Regression-Aware Inference for Language models (RAIL) framework~\cite{lukasik-etal-2024-regression}, which seeks predictions that minimize expected error under a minimum Bayes-risk decoding objective. For our discrete target space $\mathcal{Y}=\{1,2,3,4,5\}$, the expected value can be computed exactly by scoring all candidates and taking their probability-weighted average:
\begin{equation}
    \hat{y}_{m}(x) = \sum_{y \in \mathcal{Y}} 
    p(\text{str}(y)\mid x)\,y,
\end{equation}
where $\text{str}(\cdot)$ denotes the string representation of each candidate extracted from the MLLM output logits. This decision rule yields smoother, more consistent ordinal predictions than simple argmax decoding.

To align training with this inference procedure, we integrate the decision rule directly into fine-tuning via Regression-Aware Fine-Tuning (RAFT)~\cite{lukasik2025better}. Specifically, each dimension-specific adapter is optimized using mean squared error (MSE) between predicted and gold scores:
\begin{equation}
  \mathcal{L}_{m} = 
  \frac{1}{N} 
  \sum_{j=1}^{N} 
  \bigl(\hat{y}_{m}^{(j)} - y_{m}^{*\,(j)}\bigr)^{2},
\end{equation}
where $\{y_{m}^{*\,(j)}\}_{j=1}^{N}$ are the ground-truth scores for dimension $m$. This directly optimizes for regression performance while preserving the discrete nature of the targets. 
Combined with our multi-LoRA architecture, RAFT/RAIL produces a modular and interpretable system for fine-grained art assessment (Figure~\ref{fig:fig2}, right). By isolating updates within each LoRA adapter, we further reduce inter-dimensional interference and enhance specialization. Moreover, the architecture allows flexible composition: any subset of adapters $\mathcal{S} \subseteq \mathcal{D}$ can be selectively activated for context-specific evaluation or targeted feedback, mirroring how human educators emphasize different rubric criteria in different contexts.

\begin{figure*}[!ht]
  \centering
  \includegraphics[scale=0.23]{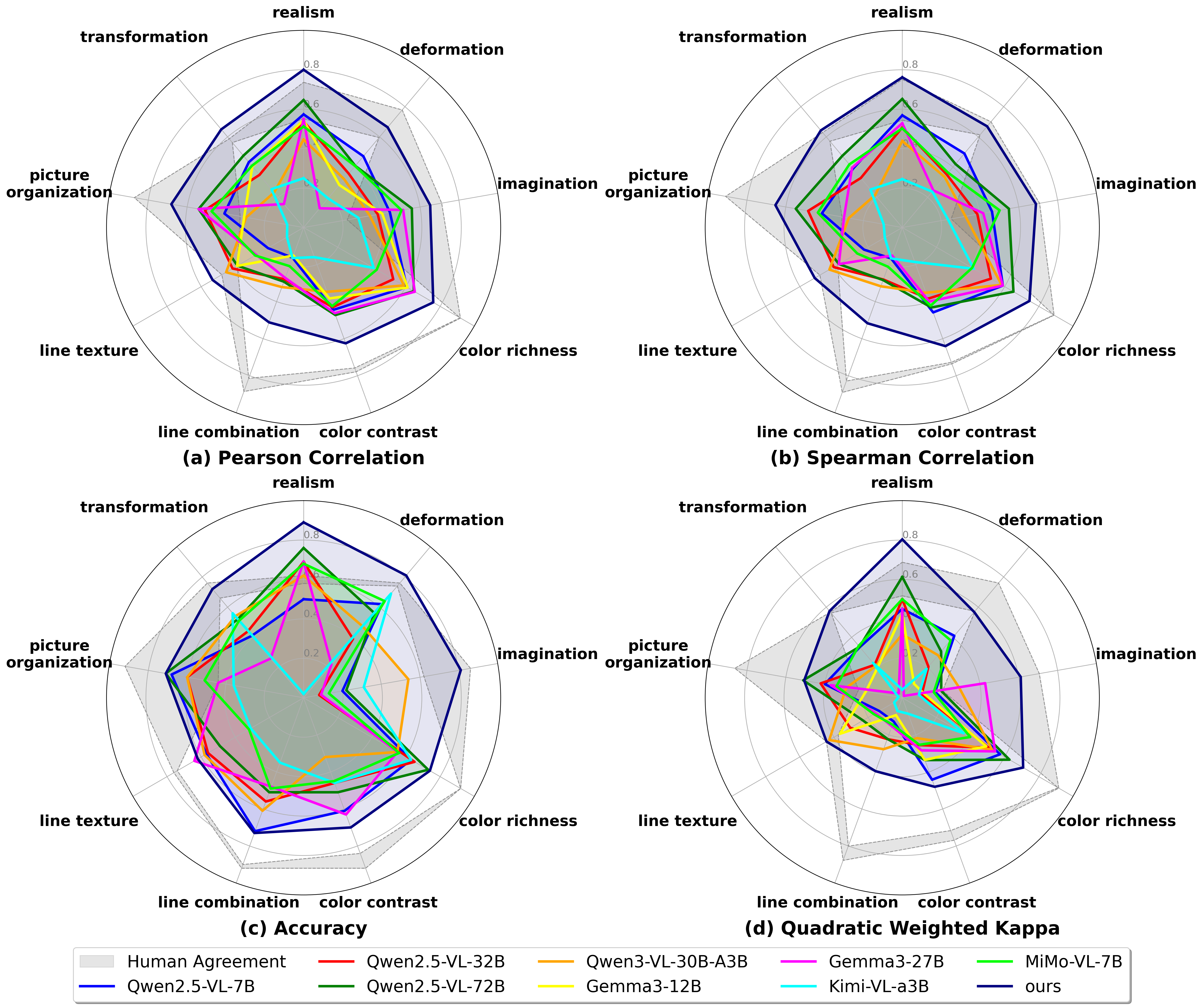}
  \caption{Radar plot showing per-dimension performance of MLLMs on KidsArtBench. We compare zero-shot prompting, our attribute-aware fine-tuned model, and human inter-rater agreement.}
  \label{fig:fig4}
\end{figure*}

\section{Results}
% \hq{do you have results comparison, using vanilla MLLMs, and after using the fine-tune LoRA. also showing the fine-grained(9-dimensional) metrics; then explain why performance on some dimensions are not good as others (perhaps can from your attention map?).}

%This section investigates the potential of MLLMs for the automated assessment of student artworks, with the focus on both performance and interpretability. We first conduct attribute-aware evaluation with top open-source MLLMs through prompting experiments. To validate the efficacy of our ArtMentor dataset and demonstrate substantial performance gains, we fine-tune Qwen2.5-VL-7B on our ArtMentor dataset employing attribute-specific multi-LoRA and RAFT framework. Additionally, we investigate and refine the training strategy by using human-written comments for guidance, and apply image augmentation to further improve robustness. Finally, we present the qualitative analysis of the multi-LoRA setting to discuss its interpretability.

\subsection{Prompting-Based Evaluation Results}\label{sec:prompt}

%We conduct prompting experiments with 8 open-sourced MLLMs: \textbf{Qwen3-VL-30B-A3B} \textit{(Oct. 2025)}, \textbf{Qwen2.5-VL-7B / 32B / 72B} \textit{(April 2025)}, \textbf{Gemma3-12B / 27B} \textit{(April 2025)}, \textbf{Mimo-VL-7B} \textit{(June 2025)}, and \textbf{Kimi-VL-A3B} \textit{(June 2025)}.
We evaluate 8 open-source MLLMs on the KidsArtBench dataset using rubric-aligned prompting: Qwen3-VL-30B-A3B (Oct. 2025), Qwen2.5-VL-7B / 32B / 72B (Apr. 2025), Gemma3-12B / 27B (Apr. 2025), Mimo-VL-7B (June 2025), and Kimi-VL-A3B (June 2025). Performance is assessed across five metrics: Spearman's rank correlation (SC), Pearson's correlation (PC), exact match accuracy (ACC), mean squared error (MSE), and quadratic weighted kappa (QWK). These collectively capture both ordinal and regression quality, as well as model-human agreement beyond chance.

A comparative performance summary of attribute-aware prompting across all models is shown in Figure~\ref{fig:fig4}, with detailed results in Tables~\ref{tab:Qwen2.5-VL-72B}-\ref{tab:Teacher2}. Among all models, the Qwen-VL family demonstrates the strongest performance. Qwen2.5-VL-72B achieves the best overall prompting results (SC = 0.487, PC = 0.492, QWK = 0.376), particularly in dimensions such as Realism (ACC = 0.76) and Picture Organization (QWK = 0.510). Interestingly, the newer Qwen3-VL-30B-A3B, while stronger on general VQA benchmarks, underperforms Qwen2.5-VL-7B in our task ($\Delta$SC = –0.105, $\Delta$QWK = –0.078), possibly due to inconsistent expert routing in its Mixture-of-Experts (MoE) architecture. Gemma3-27B shows strong performance on abstract dimensions such as Imagination (QWK = 0.557) and Color Contrast (QWK = 0.686), but struggles with others such as Deformation (SC = 0.118). Mimo-VL-7B yields a balanced overall profile, close to Qwen2.5-VL-7B. In contrast, Kimi-VL-A3B lags across all metrics, with low scores on Realism (ACC = 0.02) and an overall SC of just 0.215. 

%Across evaluated dimensions, Color Richness stands out as the most reliably assessed attribute. Qwen2.5-VL-72B achieves a SC of 0.651 and a QWK of 0.628, while Gemma3-27B attains slightly higher correlation with an SC of 0.655 and a QWK of 0.544. Realism also proves consistently more manageable, with Qwen models maintaining correlations in the range of 0.635 to 0.652. In contrast, creative and structural dimensions such as Imagination and Line Combination remain particularly challenging. For instance, performance on Line Combination rarely exceeds an SC of 0.28 or a QWK of 0.22 across all MLLMs. This disparity highlights that current MLLMs are considerably more proficient in capturing objective, surface-level perceptual qualities such as Color Richness, Color Contrast, and Realism, yet struggle to evaluate higher-order dimensions that require abstraction and interpretation, including Imagination, and the integration of lines and textures in artistic expression.
Across dimensions, Color Richness emerges as the most consistently well-predicted attribute: Qwen2.5-VL-72B achieves SC = 0.651 and QWK = 0.628, while Gemma3-27B slightly surpasses this with SC = 0.655. Realism is also reliably captured (SC $\approx$ 0.635-0.652 for Qwen models). However, more complex and abstract dimensions -- Imagination, Line Combination, Line Texture -- remain persistently challenging: performance on Line Combination, for instance, rarely exceeds SC = 0.28 or QWK = 0.22. These results indicate a key limitation of prompt-based approaches: MLLMs tend to perform better on surface-level, perceptual attributes (e.g., color, realism) but struggle with evaluative dimensions that require abstraction, compositional reasoning, or creative interpretation.

\begin{table*}[!ht]
\centering
\resizebox{\textwidth}{!}{%
\small
\begin{tabular}{ c | l | c c c c c c c c c c}
\toprule
\textbf{Metrics} & \textbf{Methods} & \textbf{Realism} & \textbf{Deformation} & \textbf{Imagination} & \textbf{Color Richness} & \textbf{Color Contrast} & \textbf{Line Combination} & \textbf{Line Texture} & \textbf{Picture Organization} & \textbf{Transformation} & \textbf{Average} \\
\midrule
& Prompting & 0.569 & 0.490 & 0.462 & 0.589 & 0.458 & 0.167 & 0.224 & 0.414 & 0.389 & 0.418 \\
& LoRA & 0.457 & 0.524 & 0.581 & 0.615 & 0.517 & 0.254 & 0.285 & 0.583 & 0.374 & 0.466 \\
\textbf{SC} {$\uparrow$}  & Multi-LoRA & 0.618 & 0.553 & 0.584 & 0.662 & 0.573 & 0.321 & 0.335 & 0.469 & 0.448 & 0.507 \\
& LoRA+RAFT & 0.545 & 0.465 & 0.567 & 0.700 & \underline{0.650} & 0.367 & 0.381 & 0.466 & 0.523 & 0.518 \\
& Multi-LoRA+RAFT & \textbf{0.762} & \underline{0.670} & \underline{0.689} & \underline{0.745} & 0.640 & \underline{0.517} & \textbf{0.513} & \underline{0.654} & \textbf{0.643} & \textbf{0.648} \\
& Human Upbound & \underline{0.751} & \textbf{0.700} & \textbf{0.707} & \textbf{0.889} & \textbf{0.736} & \textbf{0.899} & \underline{0.483} & \textbf{0.910} & \underline{0.629} & \underline{0.600} \\
\midrule
& Prompting & 0.574 & 0.472 & 0.445 & 0.604 & 0.444 & 0.163 & 0.209 & 0.408 & 0.433 & 0.417 \\
& LoRA & 0.459 & 0.518 & 0.573 & 0.617 & 0.519 & 0.245 & 0.278 & 0.612 & 0.377 & 0.466 \\
\textbf{PC} {$\uparrow$} & Multi-LoRA & 0.618 & 0.514 & 0.577 & 0.661 & 0.600 & 0.323 & 0.344 & 0.486 & 0.456 & 0.508 \\
& LoRA+RAFT & 0.541 & 0.447 & 0.569 & 0.694 & \underline{0.625} & 0.356 & 0.395 & 0.476 & 0.519 & 0.514 \\
& Multi-LoRA+RAFT & \textbf{0.800} & \underline{0.664} & \underline{0.652} & \underline{0.760} & \underline{0.625} & \underline{0.512} & \textbf{0.533} & \underline{0.682} & \textbf{0.650} & \textbf{0.653} \\
& Human Upbound & \underline{0.737} & \textbf{0.778} & \textbf{0.711} & \textbf{0.919} & \textbf{0.779} & \textbf{0.885} & \underline{0.479} & \textbf{0.872} & \underline{0.587} & \underline{0.575} \\
\midrule
& Prompting      & 0.500 & 0.620 & 0.200 & 0.630 & 0.610 & 0.720 & 0.570 & 0.680 & 0.410 & 0.549 \\
& LoRA           & 0.700 & \underline{0.770} & 0.790 & 0.660 & 0.690 & 0.650 & 0.500 & \underline{0.770} & 0.630 & 0.684 \\
\textbf{ACC} {$\uparrow$} & Multi-LoRA     & \underline{0.760} & \textbf{0.810} & 0.790 & 0.670 & 0.710 & 0.720 & 0.540 & 0.710 & 0.640 & 0.706 \\
& LoRA+RAFT      & 0.750 & 0.720 & \underline{0.810} & 0.710 & \underline{0.750} & \underline{0.760} & 0.590 & 0.720 & \underline{0.720} & 0.726 \\
& Multi-LoRA+RAFT & \textbf{0.890} & \textbf{0.810} & \underline{0.810} & \underline{0.740} & 0.700 & 0.730 & \underline{0.620} & 0.710 & \underline{0.720} & \underline{0.748} \\
& Human Upbound & 0.620 & 0.760 & \textbf{0.860} & \textbf{0.920} & \textbf{0.920} & \textbf{0.910} & \textbf{0.760} & \textbf{0.920} & \textbf{0.760} & \textbf{0.771} \\
\midrule
& Prompting & 0.560 & 0.410 & 1.520 & 0.400 & 0.390 & 0.280 & 0.430 & 0.320 & 0.710 & 0.558 \\
& LoRA & 0.330 & 0.230 & 0.210 & 0.370 & 0.340 & 0.350 & 0.530 & 0.230 & 0.370 & 0.329 \\
\textbf{MSE} {$\downarrow$} & Multi-LoRA & \underline{0.240} & \underline{0.220} & 0.210 & 0.330 & 0.290 & 0.280 & 0.490 & 0.290 & 0.360 & 0.301 \\
& LoRA+RAFT & 0.250 & 0.310 & 0.190 & 0.290 & 0.250 & 0.240 & 0.410 & 0.280 & \underline{0.280} & 0.278 \\
& Multi-LoRA+RAFT & \textbf{0.106} & \textbf{0.153} & \underline{0.170} & \underline{0.201} & \underline{0.227} & \underline{0.199} & \textbf{0.247} & \underline{0.178} & \textbf{0.237} & \textbf{0.191} \\
& Human Upbound & 0.360 & 0.240 & \textbf{0.140} & \textbf{0.080} & \textbf{0.140} & \textbf{0.100} & \underline{0.260} & \textbf{0.080} & 0.360 & \underline{0.272} \\
\midrule
& Prompting & 0.451 & 0.411 & 0.169 & 0.573 & 0.442 & 0.128 & 0.134 & 0.399 & 0.335 & 0.338 \\
& LoRA & 0.450 & 0.511 & \underline{0.610} & 0.615 & 0.519 & 0.238 & 0.270 & \underline{0.610} & 0.359 & 0.461 \\
\textbf{QWK} {$\uparrow$} & Multi-LoRA & 0.613 & 0.505 & 0.574 & 0.654 & 0.600 & 0.310 & 0.332 & 0.481 & 0.430 & 0.499 \\
& LoRA+RAFT & 0.537 & 0.429 & 0.558 & 0.693 & \underline{0.610} & 0.330 & \underline{0.385} & 0.428 & 0.497 & 0.496 \\
& Multi-LoRA+RAFT & \textbf{0.804} & \underline{0.568} & 0.573 & \underline{0.708} & 0.480 & \underline{0.396} & \textbf{0.444} & 0.505 & \textbf{0.574} & \textbf{0.566} \\
& Human Upbound & \underline{0.688} & \textbf{0.760} & \textbf{0.706} & \textbf{0.917} & \textbf{0.769} & \textbf{0.878} & \textbf{0.444} & \textbf{0.863} & \underline{0.567} & \underline{0.564} \\
\bottomrule
\end{tabular}
}
\caption{Performance of Qwen2.5-VL-7B across different training configurations. Best and second-best scores are highlighted in \textbf{bold} and \underline{underlined}, respectively. Human Upper Bound reports the better score from two human annotators per dimension. The Average cell in that row reflects the mean performance across the two human raters.} %, rather than the maximum
\label{tab:qwen_all}
\end{table*}

%%To contextualize model performance in relation to human capabilities, we invited two in-service art teachers, each with over two years of teaching experience, to annotate the test dataset, providing a human-level performance comparison. The human agreements, shown as the gray zone in Figure~\ref{fig:fig4}, is calculated by comparing teachers' scores to the gold standard, revealing a substantial performance gap when contrasted with all MLLMs. Specifically, the highest-scoring Qwen2.5-VL-72B shows a significant average QWK deficit of -0.188 against human agreement, with the most pronounced gaps in Line Combination ($\Delta$QWK = -0.618) and Color Richness ($\Delta$QWK = -0.287), meanwhile correlations and accuracy are universally lower across almost all dimensions ($\Delta$SC = 0.164, $\Delta$ACC = 0.214). Teachers exhibit extremely strong and stable agreement, particularly in evaluating color and line related dimensions, far surpassing model capabilities.

\paragraph{Prompting Strategies} We conduct ablation experiments comparing two alternative prompting setups: minimal prompting and few-shot prompting, with prompting examples and results shown in Appendix~\ref{sec:appendix_prompt} Tables~\ref{tab:simple_qwen}, \ref{tab:fewshot_qwen}. In minimal prompting, models are asked to assess each artwork based on a given dimension using a short textual instruction, without including any rubric definitions. In contrast, the few-shot prompting setup provides two exemplar image–text pairs from the training set, one low-quality (score 1) and one high-quality (score 5), each accompanied by a brief dimension-specific description to guide the model's inference. Across all dimensions, both the minimal and few-shot strategies consistently underperformed relative to the atribute-aware prompting approach. This suggests that structured rubrics offer essential guidance for model reasoning in aesthetic evaluation, while exemplar-based few-shot learning provides only limited benefit. These findings further underscore that current MLLMs, even when aided with prompting strategies, remain insufficient to achieve human-level reliability, particularly for abstract or higher-order dimensions of artistic assessment.

\paragraph{Human Agreement} To contextualize these results, two in-service art teachers (over 2 years experience) re-annotated the test set. The gray band in Figure~\ref{fig:fig4} shows inter-rater agreement. Even the best model, Qwen2.5-VL-72B, falls short of human-level performance: $\Delta$QWK = –0.188 on average, with the largest discrepancies in Line Combination ($\Delta$QWK = –0.618) and Color Richness ($\Delta$QWK = –0.287). Teachers show especially high agreement in color and line-based dimensions -- areas where MLLMs fall short. Taken together, these results demonstrate that prompting alone is insufficient for high-fidelity, human-aligned art evaluation, particularly for abstract, higher-order rubric dimensions. This motivates the need for fine-tuning approaches that better encode pedagogically relevant criteria.

%%We conducte ablation experiments using simple prompting and few-shot settings, as results shown in Tabel~\ref{tab:simple_qwen}-~\ref{tab:fewshot_qwen}. In the simple prompting setup, models receive queries in a standard scoring format without rubrics details. The few-shot setup, whereas, provide models with two exemplars retrieved from the training data, facilitating two-shot inference with rubrics guidance. Across all dimensions, both simple and few-shot prompting yielded consistently lower scores than the rubric-guided method, indicating that structured rubrics provide essential guidance for model reasoning in aesthetic evaluation, while additional exemplars alone offer limited benefit. These results collectively indicate that current MLLMs with prompting strategies are insufficient to bridge the gap to human-level reliability, especially for higher-order dimensions of artistic assessment.

\subsection{Attribute-Aware Fine-Tuning}

%Given considerations of both model performance and efficiency, we select Qwen2.5-VL-7B as our base model for subsequent fine-tuning, following the methodologies outlined in Section~\ref{sec:method}. To evaluate the effectiveness of the multi-LoRA with RAFT framework, we conduct a series of comparative experiments, as results can be found in Tabel~\ref{tab:qwen_all}. For the multi-LoRA module, we compare the proposed prompting strategy and shared-LoRA configuration to our attribute-aware settings. For RAFT, we contrast it with the standard regression-head-based method to examine RAFT contribution to feature alignment and prediction stability. 

To balance performance and efficiency, we select Qwen2.5-VL-7B as the base model for fine-tuning, following the methodology described in Section~\ref{sec:method}. To assess the effectiveness of our proposed attribute-aware multi-LoRA architecture with RAFT, we conduct a series of comparative experiments, summarized in Table~\ref{tab:qwen_all}. Specifically, we evaluate the contributions of (i) the multi-LoRA design, by comparing it against a shared-LoRA configuration, and (ii) RAFT, by comparing it against a standard regression head trained with MSE loss.

%The shared LoRA plus Regression configuration achieves moderate overall performance (SC = 0.466, QWK = 0.461). It performs relatively well on Imagination (SC = 0.581) and Picture Organization (SC = 0.583), but struggles with structural dimensions like Line Combination (SC = 0.254) and Line Texture (SC = 0.285), revealing its limitations in capturing fine-grained compositional features. Introducing a multi-LoRA design brings steady gains, raising the average SC to 0.507 and QWK to 0.499, with particular improvements in Realism and Color Contrast. Adding the RAFT architecture proves especially beneficial for perceptual attributes, where Color Richness reaches an SC of 0.700 and QWK of 0.693, clearly surpassing regression-based approaches. The combined multi-LoRA and RAFT approach demonstrates superior performance, achieving SC of 0.648, QWK of 0.566, and ACC of 0.748, and consistently outperforms all alternative configurations. The improvements are most notable in Realism, where the SC rises to 0.762 and ACC surpasses 0.890, and also in Imagination and Deformation. These findings indicate that applying our training strategies with the KidsArtBench dataset markedly strengthens MLLMs' capacity to evaluate student artwork, enhancing not only their sensitivity to perceptual fidelity but also their competence in capturing higher-level dimensions of creative abstraction. Furthermore, we perform fine-grained ablation studies and hyperparameter explorations, with more details are presented in Appendix~\ref{app:ablation}.
The shared-LoRA + standard regression baseline achieves moderate overall performance (SC = 0.466, QWK = 0.461). While it performs relatively well on dimensions such as Imagination (SC = 0.581) and Picture Organization (SC = 0.583), it struggles on structural attributes such as Line Combination (SC = 0.254) and Line Texture (SC = 0.285), highlighting its limitations in modeling fine-grained compositional features. 

Introducing multi-LoRA yields consistent improvements, increasing avg SC to 0.507 and QWK to 0.499, with notable gains in Realism and Color Contrast. Adding RAFT further enhances performance, especially on perceptual dimensions, where Color Richness reaches SC = 0.700 and QWK = 0.693, outperforming standard regression training. The full multi-LoRA + RAFT configuration yields the strongest results overall (SC = 0.648, QWK = 0.566, ACC = 0.748), outperforming all baselines across most dimensions. The largest gains appear in Realism (SC = 0.762, ACC = 0.890), as well as in Imagination and Deformation, indicating the model benefits not only from greater perceptual alignment but also from improved abstraction and compositional reasoning. These results demonstrate that our fine-tuning strategy substantially enhances MLLMs' ability to perform nuanced, multi-dimensional aesthetic assessment.

%Using the best human annotations to construct the upper-bound comparison, our method surpasses human-level performance in several dimensions, such as Realism ($\Delta$SC = 0.11, $\Delta$QWK = 0.116) and Line Texture ($\Delta$SC = 0.30). While the overall results approach human performance (average $\Delta$SC = 0.014, $\Delta$QWK = 0.007), the model still falls short of human evaluators in perceptual and stylistic dimensions, particularly those related to line and color attributes, consistent with the analysis results presented in Section~\ref{sec:prompt}.

To contextualize model performance relative to humans, we compare our best model against expert annotations. Interestingly, the model surpasses human-level agreement in select dimensions, including Realism ($\Delta$SC = +0.11, $\Delta$QWK = +0.116) and Line Texture ($\Delta$SC = +0.30). While overall model performance approaches human-level reliability (average $\Delta$SC = +0.014, $\Delta$QWK = +0.007), it still lags behind in stylistic dimensions -- particularly those involving line and color -- consistent with our earlier prompting analysis (Section~\ref{sec:prompt}). Full details on ablation settings and hyperparameter sensitivity are in Appendix~\ref{app:ablation}.

\subsection{Comments and Data Augmentation}

% In addition to score prediction, we explore whether linguistic feedback can improve model learning by using expert-written comments for auxiliary supervision and instruction-level augmentation. We restrict experiments to this subset since only these samples include narrative feedback aligned with rubric dimensions, allowing us to test if natural language comments meaningfully guide prediction -- a setting that also reflects the low-resource conditions in educational domains.

To examine how expert-written comments contribute to score assessment, we incorporate them as auxiliary linguistic feedback and evaluate their effect on both the full dataset and a 100-sample subset. We retain the attribute-specific multi-LoRA model and modify only the input $x$, which now includes additional comment text alongside the existing rubric-based instructions (prompt example with comments shown in Appendix~\ref{sec:appendix_prompt}). 

% To incorporate teacher-written comments into training, we retain the original attribute-specific multi-LoRA model and modify only the input $x$, which now includes additional comment text alongside the existing rubric-based instructions. While the comments remain fixed, we vary the instructional phrasing across prompt variants, enriching the training context without altering the model or supervision setup (prompt example with comments shown in Appendix~\ref{sec:appendix_prompt}).

On the full dataset, comment-aware training led to modest but consistent gains on several challenging dimensions, as the results shown in Table ~\ref{tab:comment_full}. Imagination exhibits substantial improvement, with PC increasing from 0.575 to 0.677. Also, line Combination shows a comparable positive shift, with PC increasing from 0.325 to 0.445, indicating that comment-based supervision is particularly helpful for dimensions that require higher-level structural reasoning. In contrast, some predominantly perceptual dimensions exhibit small declines—Realism, for instance, shows a PC drop from 0.726 to 0.697—suggesting that comments, which emphasize semantic or conceptual aspects, are more aligned with higher-level reasoning dimensions than with attributes driven primarily by low-level visual cues.

For the ablation experiments on the subset, training without comment supervision yields weak performance (avg SC of 0.304 and QWK of 0.25). Adding comments as auxiliary input raises the average correlation to 0.355, with clear improvements in dimensions such as Color Contrast ($\Delta$SC = +0.321), Realism (+0.066), and Picture Organization (+0.067). Expanding the training set through duplicating each comment-scored sample with multiple versions of instructional prompts leads to an average SC increase of 0.174, with Realism reaching SC = 0.654 and QWK = 0.485, and Imagination improving to SC = 0.531, more than doubling its baseline performance (Tables~\ref{tab:subset}–\ref{tab:comment+aug}).

We also test whether visual robustness can be improved through image augmentation, as the results shown in Table~\ref{tab:aug}. Consequently, moderate color jitter (brightness/contrast/saturation = 0.2, hue = 0.05) yields small robustness gains, whereas stronger distortions reduce performance, suggesting that maintaining perceptual fidelity—alongside curated linguistic feedback and controlled augmentation—better supports the model’s reasoning.

\subsection{Qualitative Analysis}
\label{Qual Analysis}
%We conduct qualitative investigation to better understand the behavior of our fine-tuned multi-LoRA ArtMentor base model for multi-dimension ArtMentor dataset.

%To examine the multi-LoRA settings, we analyze the subspace overlapping scores~\cite{ilharco2023editing}, which quantify the independence of different LoRA branches and reveal the trade-off between redundancy and specialization across dimensions. As shown in the subspace overlapping matrix in Figure~\ref{fig:fig5} (left), adapters interact with general low overlapping score. Notably, Deformation and Transformation exhibit an overlap of 0.328, suggesting that both capture abstract and holistic aesthetic concepts. Similarly, Line Combination and Line Texture overlap at 0.404, while Color Contrast and Color Richness reach 0.329. In both cases, these dimensions belong to the same higher-level aesthetic category, indicating that the adapters tend to focus on related visual aspects. Additionally, we compute the dimensional correlation matrix for our dataset (Figure~\ref{fig:fig5}, right). We can see that the subspace overlap and data-level correlation matrices consistently reveal that dimensions within the same higher-level aesthetic categories (e.g., Deformation–Transformation, Color Contrast–Color Richness, Line Combination–Line Texture) exhibit certain overlaps and higher correlations, while other dimensions remain largely independent, reflecting both coherent groupings and strong disentanglement across visual concepts. More qualitative analysis are shown in Appendix~\ref{app:qua}.

\begin{figure}[htbp]
  \centering
  \includegraphics[scale=0.16]{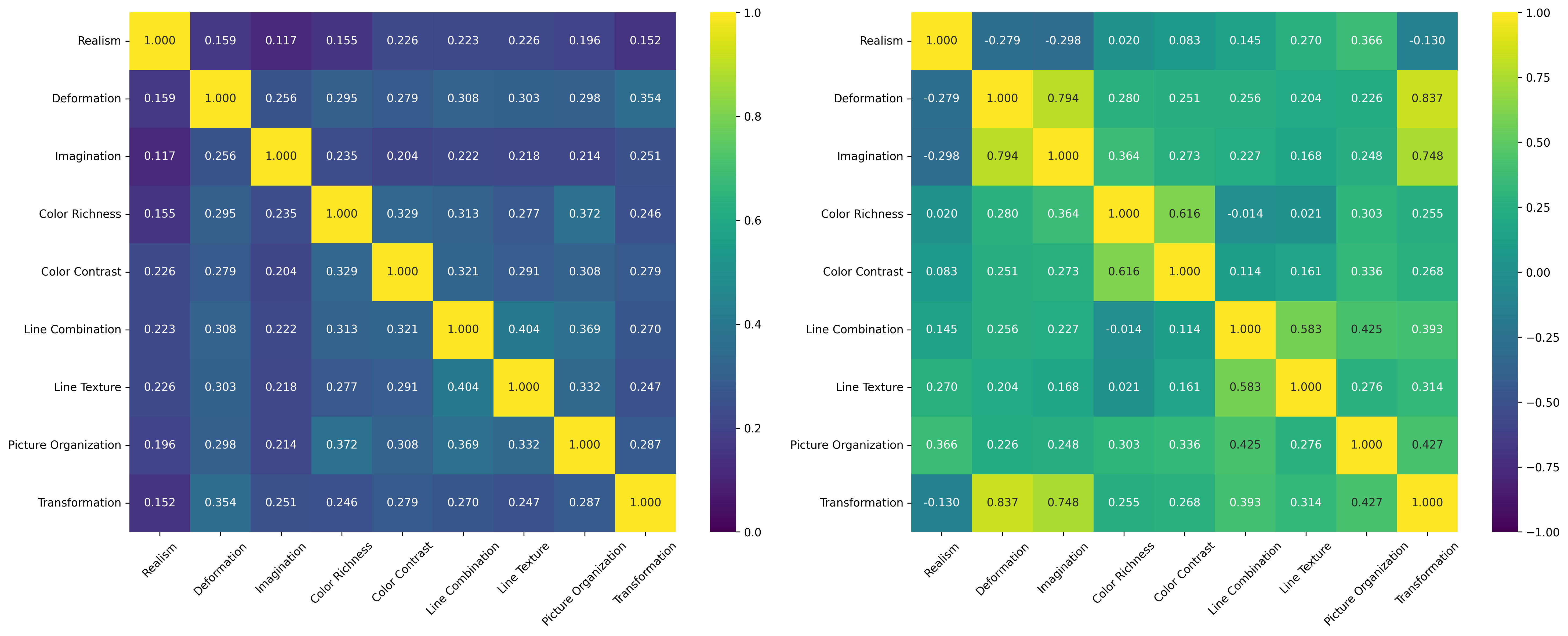}
  \caption{Subspace overlap of multi-LoRA adapters (left) and empirical dimension correlations (right).}
  %\caption{Comparative analysis for multiple dimensions with the subspace overlapping matrix for multi-LoRA settings (Left) and dimension correlation (Right).}
  \label{fig:fig5}
\end{figure}

To better understand how the model internalizes rubric dimensions, we analyze our fine-tuned multi-LoRA architecture using subspace overlapping scores~\cite{ilharco2023editing} (Figure~\ref{fig:fig5}, left), which quantify the independence of LoRA adapters—where lower overlap reflects greater specialization and higher values indicate shared representations. Notably, Deformation and Transformation show a moderate overlap of 0.328, suggesting they capture similar abstract or conceptual features. Likewise, Line Combination and Line Texture overlap at 0.404, and Color Contrast and Color Richness at 0.329. These pairs fall within shared higher-level aesthetic categories, such as line expressiveness or color dynamics, reflecting how the adapters naturally cluster around semantically related features. To compare this to real dataset distribution, we also compute the inter-dimensional correlation matrix over the annotated scores (Figure~\ref{fig:fig5}, right). The correlation patterns align closely with subspace overlap: dimensions that share visual or conceptual grounding (e.g., Deformation–Transformation, Color Contrast–Color Richness) show stronger correlations, while unrelated dimensions (e.g., Realism vs. Imagination) remain more independent.

Overall, these analyses suggest that the multi-LoRA model captures meaningful structure in the underlying rubric, with distinct yet interpretable groupings that align with educator-defined aesthetic dimensions. Additional qualitative results and more analyses are included in Appendix~\ref{app:qua}.
% \begin{figure}[!ht]
%   \centering
%   \includegraphics[scale=0.32]{fig7.png}
%   \caption{Spatial Attention Divergence between Dimensions.}
%   \label{fig:fig7}
% \end{figure}

% Furthermore, We propose to use the cross-dimensional attention divergence framework to measure the statistical distance between spatial attention distributions across dimensions, enabling quantitative comparison of visual reasoning patterns. Specificly, we calculate pairwise Jensen-Shannon distances between spatial attention distributions extracted from Peft models across different dimensional configurations. Color dimensions show maximum divergence (JS=0.54) from all others, while structural dimensions form a coherent cluster with minimal divergence (JS<0.11). This reveals a fundamental split between color processing and structural analysis in visual attention patterns.

\subsection{Error Analysis}
To better characterize model limitations, we analyzed failure cases by identifying error cases as more than two mispredicted dimensions on the test dataset. Lower-grade students’ artworks (grades 1–3, mostly ages 6–11) make up roughly 60\% of the dataset yet account for 88\% of these high-error cases, suggesting that early-grade drawings—often more abstract and structurally irregular—pose greater challenges. Error patterns vary across dimensions: Transformation and Line Texture show the highest error rates (both 61.1\%), followed by Line Combination (52.7\%), whereas perceptual attributes such as Realism exhibit much lower error rates (13.9\%). Despite these discrepancies, prediction deviations typically remain within ±1 of the ground-truth score. Media-type distributions among high-error samples (47.2\% marker, 16.7\% oil pastel, 19.4\% crayon/colored pencil, 11.1\% watercolor) largely mirror their overall dataset frequencies, indicating no strong medium-specific effects. This further shows that failures commonly arise in drawings that are highly schematic, contain overlapping or fragmented objects, or employ unconventional spatial layouts.

\section{Conclusions}

We present KidsArtBench, the first public benchmark for multi-dimensional evaluation of children's artwork using rubric-guided annotations. %The dataset includes 1,046 student submissions, scored by expert educators across 9 formative dimensions, with a subset featuring bilingual feedback comments. 
To model this structured aesthetic assessment task, we introduce an attribute-aware fine-tuning framework combining Multi-LoRA (with one adapter per rubric dimension) and RAFT, aligned with regression-aware inference.
Our method significantly outperforms prompting and shared-adapter baselines, achieving strong agreement with expert ratings. Analyses of adapter subspaces and attention patterns confirm that the model learns distinct, interpretable representations for each attribute. Further improvements from comment supervision and instruction-level augmentation highlight the value of linguistic and pedagogical cues. KidsArtBench bridges multimodal AI and education, demonstrating that properly aligned MLLMs can deliver nuanced, rubric-based assessment in creative domains. We release all data, code, and annotations to support future research in educationally grounded multimodal learning.

\section*{Limitations}

%This study has several limitations that open directions for future work. Our benchmark is currently under expansion, as only part of the dataset includes comment annotations; we plan to extend both its scale and annotation coverage to support broader evaluation. In addition, comments are used in this work merely as auxiliary training signals. Future research will explore leveraging them as intermediate supervision for reinforcement learning or chain-of-thought–based assessment, and as nature language generation targets for pedagogically oriented feedback. Also, constrained by computational resources, our fine-tuning experiments focus on a 7B-scale model. With advances in efficiency such as MoE models, future work will investigate scaling to larger and more capable MLLMs.
This study has several limitations that suggest directions for future work. Our dataset only contain about 1k samples and focuses primarily on children’s artworks from a specific region. Despite many efforts we made for generalization, we fully acknowledge the demographic and cultural limitations of the current dataset. Conducting large-scale cross-cultural data collection lies beyond the scope of the present work, but we view it as an important direction for future work, including consideration of potential extensions such as more systematic cross-regional sampling and explicit cultural comparisons. Additionally, comments are used solely as auxiliary training signals in this paper. Future research will explore their use as intermediate supervision for reinforcement learning, chain-of-thought–style assessment, and as targets for generating pedagogically grounded feedback. Due to computational constraints, our fine-tuning experiments focus on a 7B-scale model. As efficiency improves -- e.g., with Mixture-of-Experts (MoE) architectures -- future work will explore scaling to larger and more capable MLLMs.

\section*{Ethics Statement and Bias}

KidsArtBench currently contains children artwork (aged 5 to 15) collected from multiple primary and middle schools predominantly located in Eastern China, as well as several summer-camp programs whose participants come from diverse regions across the country. This dataset is  under protocol WZU-2025-106 approved by the East China Normal University Institutional Review Board on 20 August 2025, in line with the Personal Information Protection Law of the People's Republic of China (PIPL) where applicable and established academic ethical policies. We obtained parent/guardian consent and child assent; forms specified collected data (artwork image, optional title, statement, age band), research purpose, security, withdrawal rights, and release conditions. Inclusion required original works by 5–15-year-olds from educational settings; we excluded items with identifying marks, identifiable photos, or missing consent. Prior to any access or release we removed direct identifiers, replaced age with bands, manually redacted signatures, assigned non-reversible IDs, and screened statements.
% ; two researchers verified de-identification (third-party tie-breaker). Raw files are held on encrypted, access-controlled servers (audit-logged) for 5 years to enable withdrawals via XXX (hidden for review); 
%51284118014@stu.ecnu.edu.cn;
The public package contains only de-identified images and labels. Annotations were produced by 12 art educators with more than 5 years' experience after an average 24-hour calibration. The dataset primarily consists of artworks collected from children across multiple primary schools and middle schools within eastern China, which may introduce regional and cultural biases inherent to the local educational context. The dataset may also reflect socioeconomic biases, as participating schools tend to represent regions with relatively higher access to art education resources.
%Our bias audit reports metrics stratified by age band, presenting gender (if provided/consented), and region/school type, includes leave-group-out tests, rater-disagreement analyses, and a qualitative safety review of model comments.
Intended use is research and formative feedback (not grading or placement). We will release our code and data (\url{https://github.com/bigrayss/KidsArtBench}) 
%(https://artmentor.github.io/)
under the ACL Code of Ethics and the MIT License with terms forbidding re-identification, high-stakes use, and unauthorised redistribution.

\bibliography{custom}

\onecolumn
\appendix
\clearpage
\section{Appendix}
\label{sec:appendix}

\subsection{Evaluation Rubrics for Prompt-based Score Assessment} \label{sec:appendix rubrics}

% This section details the evaluation criteria and scoring rubrics for each dimension for our ArtMentor dataset. Each dimension is scored on a 5-point scale.

\begin{table*}[!ht]
\centering
\small
\begin{tabular}{lp{14.5cm}}
\toprule
\textbf{Score} & \textbf{Description} \\
\midrule
5 & The artwork exhibits exceptional detail and precision in depicting Realism features. Textures and lighting are used masterfully to mimic real-life appearances with accurate proportions and perspective. The representation is strikingly lifelike, demonstrating advanced skills in realism. \\
4 & The artwork presents a high level of detail and accuracy in the portrayal of subjects. Proportions and textures are very well executed, and the lighting enhances the realism. Although highly Realism, minor discrepancies in perspective or detail might be noticeable. \\
3 & The artwork represents subjects with a moderate level of realism. Basic proportions are correct, and some textures and lighting effects are used to enhance realism. However, the depiction may lack depth or detail in certain areas. \\
2 & The artwork attempts realism but struggles with accurate proportions and detailed textures. Lighting and perspective may be inconsistently applied, resulting in a less convincing depiction. \\
1 & The artwork shows minimal attention to Realism details. Proportions, textures, and lighting are poorly executed, making the depiction far from lifelike. \\
\bottomrule
\end{tabular}
\caption{Realism assessment rubrics for prompting. This criterion assesses the accuracy of proportions, textures, lighting, and perspective to create a lifelike depiction.}
\label{tab:realism_rubric}
\end{table*}

\vspace{-0.5cm}

\begin{table*}[!ht]
\centering
\small
\begin{tabular}{lp{14.5cm}}
\toprule
\textbf{Score} & \textbf{Description} \\
\midrule
5 & The artwork demonstrates masterful use of deformation to enhance the emotional or conceptual impact of the piece. The transformations are thoughtful and integral to the artwork's message, seamlessly blending with the composition to engage viewers profoundly. \\
4 & The artwork effectively uses deformation to express artistic intentions. The modifications are well-integrated and contribute significantly to the viewer's understanding or emotional response. Minor elements of the deformation might detract from its overall effectiveness. \\
3 & The artwork includes noticeable deformations that add to its artistic expression. While these elements generally support the artwork's theme, they may be somewhat disjointed from the composition, offering mixed impact on the viewer. \\
2 & The artwork attempts to use deformation but does so with limited success. The deformations are present but feel forced or superficial, only marginally contributing to the artwork's expressive goals. \\
1 & The artwork features minimal or ineffective deformation, with little to no enhancement of the artwork's message or emotional impact. The attempts at deformation seem disconnected from the artwork's overall intent. \\
\bottomrule
\end{tabular}
\caption{Deformation assessment rubrics for prompting. This criterion evaluates the artist's ability to creatively and intentionally deform reality to convey a message, emotion, or concept.}
\label{tab:deformation_rubric}
\end{table*}

\vspace{-0.5cm}

\begin{table*}[!ht]
\centering
\small
\begin{tabular}{lp{14.5cm}}
\toprule
\textbf{Score} & \textbf{Description} \\
\midrule
5 & The artwork masterfully employs contrasting colors to create a striking and effective visual impact. \\
4 & The artwork effectively uses contrasting colors to enhance visual interest, though the contrast may be less pronounced. \\
3 & The artwork has some contrast in colors, but it is not used effectively to enhance the artwork's overall appeal. \\
2 & The artwork makes minimal use of color contrast, resulting in a lackluster visual impact. \\
1 & The artwork lacks effective color contrast, making the piece visually unengaging. \\
\bottomrule
\end{tabular}
\caption{Color contrast assessment Rubrics for prompting. This criterion evaluates the effective use of contrasting colors to enhance artistic expression.}
\label{tab:color_contrast_rubric}
\end{table*}

\vspace{-0.5cm}

\begin{table*}[!ht]
\centering
\small
\begin{tabular}{lp{14.5cm}}
\toprule
\textbf{Score} & \textbf{Description} \\
\midrule
5 & The artwork uses a wide and harmonious range of colors, each contributing to a vivid and dynamic composition. \\
4 & The artwork features a good variety of colors that are well-balanced, enhancing the visual appeal of the piece. \\
3 & The artwork includes a moderate range of colors, but the palette may not fully enhance the subject matter. \\
2 & The artwork has limited color variety, with a palette that does not significantly contribute to the piece's impact. \\
1 & The artwork shows poor use of colors, with a very restricted range that detracts from the visual experience. \\
\bottomrule
\end{tabular}
\caption{Color richness assessment rubrics for prompting. This criterion assesses the use and range of colors to create a visually engaging experience.}
\label{tab:color_richness_rubric}
\end{table*}

\vspace{-0.5cm}

\begin{table*}[!ht]
\centering
\small
\begin{tabular}{lp{14.5cm}}
\toprule
\textbf{Score} & \textbf{Description} \\
\midrule
5 & The artwork demonstrates a wide variety of line textures, each skillfully executed to enhance the piece's aesthetic and thematic elements. \\
4 & The artwork includes a good range of line textures, well executed but with some areas that may lack definition. \\
3 & The artwork features moderate variety in line textures, with generally adequate execution but lacking in detail. \\
2 & The artwork has limited line textures, with execution that does not significantly contribute to the artwork's quality. \\
1 & The artwork lacks variety and sophistication in line textures, resulting in a visually dull piece. \\
\bottomrule
\end{tabular}
\caption{Line texture assessment rubrics for prompting. This criterion evaluates the variety and execution of line textures within the artwork.}
\label{tab:line_texture_rubric}
\end{table*}

\vspace{-0.5cm}

\begin{table*}[!ht]
\centering
\small
\begin{tabular}{lp{14.5cm}}
\toprule
\textbf{Score} & \textbf{Description} \\
\midrule
5 & The artwork displays a profound level of originality and creativity, introducing unique concepts or interpretations that are both surprising and thought-provoking. \\
4 & The artwork presents creative ideas that are both original and nicely executed, though they may be similar to conventional themes. \\
3 & The artwork shows some creative ideas, but they are somewhat predictable and do not stray far from traditional approaches. \\
2 & The artwork has minimal creative elements, with ideas that are largely derivative and lack originality. \\
1 & The artwork lacks imagination, with no discernible original ideas or creative concepts. \\
\bottomrule
\end{tabular}
\caption{Imagination assessment rubrics for prompting. This criterion evaluates the artist's ability to use their creativity to form unique and original ideas within their artwork.}
\label{tab:imagination_rubric}
\end{table*}

\vspace{-0.5cm}

\begin{table*}[!ht]
\centering
\small
\begin{tabular}{lp{14.5cm}}
\toprule
\textbf{Score} & \textbf{Description} \\
\midrule
5 & The artwork exhibits exceptional integration of line combinations, creating a harmonious and engaging visual flow. \\
4 & The artwork displays good use of line combinations that contribute to the overall composition, though some areas may lack cohesion. \\
3 & The artwork shows average use of line combinations, with some effective sections but overall lacking in cohesiveness. \\
2 & The artwork has minimal effective use of line combinations, with lines that often clash or do not contribute to a unified composition. \\
1 & The artwork shows poor integration of lines, with combinations that disrupt the visual harmony of the piece. \\
\bottomrule
\end{tabular}
\caption{Line combination assessment Rubrics for prompting. This criterion assesses the integration and interaction of lines within the artwork.}
\label{tab:line_combination_rubric}
\end{table*}

\vspace{-0.5cm}

\begin{table*}[!ht]
\centering
\small
\begin{tabular}{lp{14.5cm}}
\toprule
\textbf{Score} & \textbf{Description} \\
\midrule
5 & The artwork is impeccably organized, with each element thoughtfully placed to create a balanced and compelling composition. \\
4 & The artwork has a good organization, with a well-arranged composition that effectively guides the viewer's eye, though minor elements may disrupt the flow. \\
3 & The artwork has an adequate organization, but the composition may feel somewhat unbalanced or disjointed. \\
2 & The artwork shows poor organization, with a composition that lacks coherence and does not effectively engage the viewer. \\
1 & The artwork is poorly organized, with a chaotic composition that detracts from the piece's overall impact. \\
\bottomrule
\end{tabular}
\caption{Picture organization assessment rubrics for prompting. This criterion evaluates the overall composition and spatial arrangement within the artwork.}
\label{tab:picture_organization_rubric}
\end{table*}

\vspace{-0.5cm}

\begin{table*}[!ht]
\centering
\small
\begin{tabular}{lp{14.5cm}}
\toprule
\textbf{Score} & \textbf{Description} \\
\midrule
5 & The artwork is transformative, offering a fresh and innovative take on traditional elements, significantly enhancing the viewer's experience. \\
4 & The artwork successfully transforms familiar elements, providing a new perspective, though the innovation may not be striking. \\
3 & The artwork shows some transformation of familiar elements, but the changes are somewhat predictable and not highly innovative. \\
2 & The artwork attempts transformation but achieves only minimal success, with changes that are either too subtle or not effectively executed. \\
1 & The artwork lacks transformation, with traditional elements that are replicated without any significant innovation or creative reinterpretation. \\
\bottomrule
\end{tabular}
\caption{Transformation assessment rubrics for prompting. This criterion assesses the artist's ability to transform traditional or familiar elements into something new and unexpected.}
\label{tab:transformation_rubric}
\end{table*}

\clearpage
\subsection{Prompt Example}
\label{sec:appendix_prompt}

\begin{tcolorbox}[title=\textbf{Simple Prompt Example}]
<image> Assess the student's artwork based on the 'Color contrast ' criterion and provide a score.
This criterion evaluates the effective use of contrasting colors to enhance artistic expression.
Output a score (1-5).
\end{tcolorbox}

\begin{tcolorbox}[title=\textbf{Rubric-based Prompt Example}]
<image> Assess the student's artwork based on the 'Color contrast ' criterion and provide a score.
This criterion evaluates the effective use of contrasting colors to enhance artistic expression.

5: The artwork masterfully employs contrasting colors to create a striking and effective visual impact.

4: The artwork effectively uses contrasting colors to enhance visual interest, though the contrast may be less pronounced.

3: The artwork has some contrast in colors, but it is not used effectively to enhance the artwork's overall appeal.

2: The artwork makes minimal use of color contrast, resulting in a lackluster visual impact.

1: The artwork lacks effective color contrast, making the piece visually unengaging.

Output a score (1-5).
\end{tcolorbox}

\begin{tcolorbox}[title=\textbf{Few-shot Prompt Example}]
<image> Assess the student's artwork based on the 'Color contrast' criterion and provide a score.

<image> This is an example of 'Color contrast' with a score of 1.

<image> This is an example of 'Color contrast' with a score of 5.

This criterion evaluates the effective use of contrasting colors to enhance artistic expression.

5: The artwork masterfully employs contrasting colors to create a striking and effective visual impact.

4: The artwork effectively uses contrasting colors to enhance visual interest, though the contrast may be less pronounced.

3: The artwork has some contrast in colors, but it is not used effectively to enhance the artwork's overall appeal.

2: The artwork makes minimal use of color contrast, resulting in a lackluster visual impact.

1: The artwork lacks effective color contrast, making the piece visually unengaging.

Output a score (1-5).
\end{tcolorbox}

\begin{tcolorbox}[title=\textbf{Prompt Example with Comments for Fine-tuning}]
<image> Assess the student's artwork based on the 'Color contrast ' criterion and provide a score.
This criterion evaluates the effective use of contrasting colors to enhance artistic expression.

5: The artwork masterfully employs contrasting colors to create a striking and effective visual impact.

4: The artwork effectively uses contrasting colors to enhance visual interest, though the contrast may be less pronounced.

3: The artwork has some contrast in colors, but it is not used effectively to enhance the artwork's overall appeal.

2: The artwork makes minimal use of color contrast, resulting in a lackluster visual impact.

1: The artwork lacks effective color contrast, making the piece visually unengaging.

Reference/Expert/Teacher comment: [Insert the expert’s qualitative feedback here, if available.]

Output a score (1-5).
\end{tcolorbox}

\clearpage
\twocolumn
\subsection{Detailed Results}

This section details results on our test dataset. To ensure a fair comparison across MLLMs, all images are uniformly resized to 448×448 during preprocessing, while MLLMs are configured with consistent parameters ($max\_new\_tokens = 128$, $temperature = 0.7$, $top_k = 50$, and $top_p = 1.0$). Table~\ref{tab:Qwen2.5-VL-72B}-\ref{tab:Kimi-a3b} summarize the results of our prompting experiments for attribute-aware evaluation using MLLMs. Table~\ref{tab:Teacher1}-\ref{tab:Teacher2} present the scores from two practicing art teachers, which serve as the human agreements on the test dataset. We used up to 4 NVIDIA A100 GPUs for all experiments.

\begin{table}[!ht]
\centering
\resizebox{0.9\linewidth}{!}{
\small
\begin{tabular}{lcccccc}
\toprule
\textbf{Dimension} & \textbf{SC} & \textbf{PC} & \textbf{ACC} & \textbf{MSE} & \textbf{QWK} \\
\midrule
Realism & 0.652 & 0.647 & 0.760 & 0.240 & 0.613 \\
Deformation & 0.420 & 0.408 & 0.560 & 0.470 & 0.307 \\
Imagination & 0.549 & 0.558 & 0.220 & 0.780 & 0.201 \\
Color Richness & 0.651 & 0.646 & 0.730 & 0.270 & 0.628 \\
Color Contrast & 0.432 & 0.473 & 0.510 & 0.490 & 0.336 \\
Line Combination & 0.284 & 0.292 & 0.510 & 0.490 & 0.222 \\
Line Texture & 0.371 & 0.388 & 0.490 & 0.510 & 0.230 \\
Picture Organization & 0.549 & 0.541 & 0.710 & 0.290 & 0.510 \\
Transformation & 0.474 & 0.476 & 0.520 & 0.540 & 0.338 \\
Average & 0.487 & 0.492 & 0.557 & 0.453 & 0.376 \\
\bottomrule
\end{tabular}
}
\caption{Prompting results with Qwen2.5-VL-72B}
\label{tab:Qwen2.5-VL-72B}
\end{table}

\vspace{-0.4cm}

\begin{table}[!ht]
\centering
\resizebox{0.9\linewidth}{!}{
\small
\begin{tabular}{lcccccc}
\toprule
\textbf{Dimension} & \textbf{SC} & \textbf{PC} & \textbf{ACC} & \textbf{MSE} & \textbf{QWK} \\
\midrule
Realism & 0.580 & 0.584 & 0.53 & 0.53 & 0.47 \\
Deformation & 0.490 & 0.472 & 0.62 & 0.41 & 0.411 \\
Imagination & 0.476 & 0.459 & 0.22 & 1.47 & 0.183 \\
Color Richness & 0.594 & 0.608 & 0.64 & 0.39 & 0.579 \\
Color Contrast & 0.476 & 0.459 & 0.62 & 0.38 & 0.458 \\
Line Combination & 0.167 & 0.163 & 0.72 & 0.28 & 0.128 \\
Line Texture & 0.312 & 0.288 & 0.60 & 0.40 & 0.211 \\
Picture Organization & 0.430 & 0.422 & 0.69 & 0.31 & 0.413 \\
Transformation & 0.399 & 0.444 & 0.44 & 0.68 & 0.347 \\
Overall & 0.436 & 0.433 & 0.564 & 0.539 & 0.355 \\
\bottomrule
\end{tabular}
}
\caption{Prompting results with Qwen2.5-VL-32B}
\label{tab:Qwen2.5-VL-32B}
\end{table}

\vspace{-0.4cm}

\begin{table}[!ht]
\centering
\resizebox{0.9\linewidth}{!}{
\small
\begin{tabular}{lcccccc}
\toprule
\textbf{Dimension} & \textbf{SC} & \textbf{PC} & \textbf{ACC} & \textbf{MSE} & \textbf{QWK} \\
\midrule
Realism & 0.635 & 0.637 & 0.570 & 0.460 & 0.511 \\
Deformation & 0.506 & 0.530 & 0.620 & 0.380 & 0.458 \\
Imagination & 0.495 & 0.481 & 0.200 & 1.220 & 0.194 \\
Color Richness & 0.589 & 0.604 & 0.630 & 0.400 & 0.547 \\
Color Contrast & 0.524 & 0.505 & 0.670 & 0.330 & 0.494 \\
Line Combination & 0.310 & 0.310 & 0.730 & 0.270 & 0.289 \\
Line Texture & 0.147 & 0.196 & 0.530 & 0.500 & 0.127 \\
Picture Organization & 0.579 & 0.565 & 0.750 & 0.250 & 0.550 \\
Transformation & 0.423 & 0.451 & 0.480 & 0.610 & 0.374 \\
Average & 0.468 & 0.475 & 0.576 & 0.491 & 0.394 \\
\bottomrule
\end{tabular}
}
\caption{Prompting results with Qwen2.5-VL-7B}
\label{tab:Qwen2.5-VL-7B}
\end{table}

\vspace{-0.4cm}

\begin{table}[!ht]
\centering
\resizebox{0.9\linewidth}{!}{
\small
\begin{tabular}{lcccccc}
\toprule
\textbf{Dimension} & \textbf{SC} & \textbf{PC} & \textbf{ACC} & \textbf{MSE} & \textbf{QWK} \\
\midrule
Realism & 0.439 & 0.446 & 0.620 & 0.380 & 0.321 \\
Deformation & 0.335 & 0.320 & 0.470 & 0.560 & 0.282 \\
Imagination & 0.325 & 0.342 & 0.540 & 0.520 & 0.293 \\
Color Richness & 0.582 & 0.588 & 0.550 & 0.510 & 0.527 \\
Color Contrast & 0.352 & 0.347 & 0.320 & 1.090 & 0.218 \\
Line Combination & 0.317 & 0.322 & 0.610 & 0.420 & 0.278 \\
Line Texture & 0.427 & 0.454 & 0.580 & 0.420 & 0.428 \\
Picture Organization & 0.276 & 0.296 & 0.600 & 0.430 & 0.289 \\
Transformation & 0.214 & 0.232 & 0.540 & 0.610 & 0.210 \\
Average & 0.363 & 0.372 & 0.537 & 0.549 & 0.316 \\
\bottomrule
\end{tabular}
}
\caption{Prompting results with Qwen3-VL-30B-A3B}
\label{tab:Qwen3-VL-30B-A3B}
\end{table}

\begin{table}[!ht]
\centering
\resizebox{0.9\linewidth}{!}{
\small
\begin{tabular}{lcccccc}
\toprule
\textbf{Dimension} & \textbf{SC} & \textbf{PC} & \textbf{ACC} & \textbf{MSE} & \textbf{QWK} \\
\midrule
Realism & 0.529 & 0.544 & 0.69 & 0.31 & 0.426 \\
Deformation & 0.246 & 0.282 & 0.22 & 1.05 & 0.093 \\
Imagination & 0.417 & 0.405 & 0.09 & 1.72 & 0.113 \\
Color Richness & 0.589 & 0.609 & 0.55 & 0.45 & 0.492 \\
Color Contrast & 0.400 & 0.383 & 0.63 & 0.43 & 0.336 \\
Line Combination & 0.151 & 0.151 & 0.48 & 0.67 & 0.094 \\
Line Texture & 0.371 & 0.390 & 0.64 & 0.36 & 0.366 \\
Picture Organization & 0.300 & 0.306 & 0.44 & 0.68 & 0.187 \\
Transformation & 0.396 & 0.405 & 0.26 & 1.13 & 0.197 \\
Average & 0.378 & 0.386 & 0.444 & 0.756 & 0.256 \\
\bottomrule
\end{tabular}
}
\caption{Prompting results with by Gemma3-12B}
\label{tab:Gemma3-12B}
\end{table}

\begin{table}[!ht]
\centering
\resizebox{0.9\linewidth}{!}{
\small
\begin{tabular}{lcccccc}
\toprule
\textbf{Dimension} & \textbf{SC} & \textbf{PC} & \textbf{ACC} & \textbf{MSE} & \textbf{QWK} \\
\midrule
Realism & 0.543 & 0.550 & 0.720 & 0.529 & 0.451 \\
Deformation & 0.118 & 0.127 & 0.210 & 1.127 & 0.012 \\
Imagination & 0.517 & 0.513 & 0.690 & 0.557 & 0.427 \\
Color Richness & 0.655 & 0.652 & 0.680 & 0.566 & 0.544 \\
Color Contrast & 0.447 & 0.462 & 0.560 & 0.686 & 0.284 \\
Line Combination & 0.282 & 0.272 & 0.750 & 0.500 & 0.154 \\
Line Texture & 0.215 & 0.277 & 0.600 & 0.678 & 0.156 \\
Picture Organization & 0.536 & 0.542 & 0.650 & 0.592 & 0.368 \\
Transformation & 0.147 & 0.155 & 0.310 & 0.949 & 0.029 \\
Average & 0.384 & 0.395 & 0.574 & 0.687 & 0.269 \\
\bottomrule
\end{tabular}
}
\caption{Prompting results with by Gemma3-27B}
\label{tab:Gemma3-27B}
\end{table}

\begin{table}[!ht]
\centering
\resizebox{0.9\linewidth}{!}{
\small
\begin{tabular}{lcccccc}
\toprule
\textbf{Dimension} & \textbf{SC} & \textbf{PC} & \textbf{ACC} & \textbf{MSE} & \textbf{QWK} \\
\midrule
Realism & 0.503 & 0.513 & 0.680 & 0.320 & 0.500 \\
Deformation & 0.362 & 0.407 & 0.640 & 0.420 & 0.381 \\
Imagination & 0.502 & 0.504 & 0.130 & 1.140 & 0.162 \\
Color Richness & 0.409 & 0.428 & 0.560 & 0.440 & 0.398 \\
Color Contrast & 0.422 & 0.425 & 0.450 & 0.640 & 0.254 \\
Line Combination & 0.212 & 0.208 & 0.490 & 0.570 & 0.145 \\
Line Texture & 0.264 & 0.284 & 0.320 & 1.190 & 0.166 \\
Picture Organization & 0.435 & 0.474 & 0.510 & 0.490 & 0.339 \\
Transformation & 0.418 & 0.408 & 0.510 & 0.630 & 0.344 \\
Average & 0.392 & 0.406 & 0.477 & 0.649 & 0.299 \\
\bottomrule
\end{tabular}
}
\caption{Prompting results with by Mimo-VL-7B}
\label{tab:Mimo-VL-7B}
\end{table}

\begin{table}[!ht]
\centering
\resizebox{0.9\linewidth}{!}{
\small
\begin{tabular}{lcccccc}
\toprule
\textbf{Dimension} & \textbf{SC} & \textbf{PC} & \textbf{ACC} & \textbf{MSE} & \textbf{QWK} \\
\midrule
Realism & 0.245 & 0.250 & 0.02 & 3.05 & 0.041 \\
Deformation & 0.237 & 0.193 & 0.69 & 0.40 & 0.192 \\
Imagination & 0.236 & 0.282 & 0.31 & 0.72 & 0.085 \\
Color Richness & 0.419 & 0.411 & 0.63 & 0.40 & 0.360 \\
Color Contrast & 0.186 & 0.160 & 0.46 & 0.68 & 0.086 \\
Line Combination & 0.167 & 0.165 & 0.35 & 0.74 & 0.072 \\
Line Texture & 0.102 & 0.097 & 0.31 & 1.14 & 0.048 \\
Picture Organization & 0.096 & 0.084 & 0.36 & 0.82 & 0.019 \\
Transformation & 0.251 & 0.254 & 0.56 & 0.47 & 0.232 \\
Average & 0.215 & 0.211 & 0.41 & 0.936 & 0.126 \\
\bottomrule
\end{tabular}
}
\caption{Prompting results with by Kimi-a3b}
\label{tab:Kimi-a3b}
\end{table}

\begin{table}[H]
\centering
\resizebox{0.9\linewidth}{!}{
\small
\begin{tabular}{lcccccc}
\toprule
\textbf{Dimension} & \textbf{SC} & \textbf{PC} & \textbf{ACC} & \textbf{MSE} & \textbf{QWK} \\
\midrule
Realism & 0.544 & 0.551 & 0.620 & 0.56 & 0.517 \\
Deformation & 0.700 & 0.778 & 0.760 & 0.24 & 0.760 \\
Imagination & 0.707 & 0.711 & 0.860 & 0.14 & 0.706 \\
Color Richness & 0.889 & 0.919 & 0.920 & 0.08 & 0.917 \\
Color Contrast & 0.728 & 0.758 & 0.840 & 0.22 & 0.717 \\
Line Combination & 0.828 & 0.814 & 0.920 & 0.14 & 0.801 \\
Line Texture & 0.378 & 0.389 & 0.760 & 0.30 & 0.375 \\
Picture Organization & 0.310 & 0.327 & 0.440 & 0.68 & 0.226 \\
Transformation & 0.629 & 0.587 & 0.660 & 0.46 & 0.567 \\
Average & 0.629 & 0.587 & 0.753 & 0.313 & 0.567 \\
\bottomrule
\end{tabular}
}
\caption{Teacher 1 performance}
\label{tab:Teacher1}
\end{table}

\begin{table}[H]
\centering
\resizebox{0.9\linewidth}{!}{
\small
\begin{tabular}{lcccccc}
\toprule
\textbf{Dimension} & \textbf{SC} & \textbf{PC} & \textbf{ACC} & \textbf{MSE} & \textbf{QWK} \\
\midrule
Realism & 0.751 & 0.737 & 0.580 & 0.60 & 0.688 \\
Deformation & 0.614 & 0.596 & 0.740 & 0.32 & 0.573 \\
Imagination & 0.252 & 0.225 & 0.620 & 0.50 & 0.208 \\
Color Richness & 0.888 & 0.917 & 0.920 & 0.08 & 0.913 \\
Color Contrast & 0.736 & 0.779 & 0.920 & 0.14 & 0.769 \\
Line Combination & 0.899 & 0.885 & 0.900 & 0.10 & 0.878 \\
Line Texture & 0.483 & 0.479 & 0.740 & 0.26 & 0.444 \\
Picture Organization & 0.910 & 0.872 & 0.920 & 0.08 & 0.863 \\
Transformation & 0.571 & 0.563 & 0.760 & 0.36 & 0.561 \\
Average & 0.571 & 0.563 & 0.789 & 0.271 & 0.561 \\
\bottomrule
\end{tabular}
}
\caption{Teacher 2 performance}
\label{tab:Teacher2}
\end{table}

% \begin{table}[H]
% \centering
% \resizebox{0.9\linewidth}{!}{
% \small
% \begin{tabular}{lcccccc}
% \toprule
% \textbf{Dimension} & \textbf{SC} & \textbf{PC} & \textbf{ACC} & \textbf{MSE} & \textbf{QWK} \\
% \midrule
% Realism & 0.648 & 0.644 & 0.600 & 0.580 & 0.603 \\
% Deformation & 0.657 & 0.687 & 0.750 & 0.280 & 0.667 \\
% Imagination & 0.480 & 0.468 & 0.740 & 0.320 & 0.457 \\
% Color Richness & 0.889 & 0.918 & 0.920 & 0.080 & 0.915 \\
% Color Contrast & 0.732 & 0.769 & 0.880 & 0.180 & 0.743 \\
% Line Combination & 0.864 & 0.850 & 0.910 & 0.120 & 0.840 \\
% Line Texture & 0.431 & 0.434 & 0.750 & 0.280 & 0.410 \\
% Picture Organization & 0.610 & 0.600 & 0.680 & 0.380 & 0.545 \\
% Transformation & 0.600 & 0.575 & 0.710 & 0.410 & 0.564 \\
% Average & 0.600 & 0.575 & 0.771 & 0.292 & 0.564 \\
% \bottomrule
% \end{tabular}
% }
% \caption{Average Performance of Two Teachers}
% \label{tab:AverageTeachers}
% \end{table}

\subsection{Results and Ablation for Base Model}
\label{app:ablation}

This appendix presents detailed ablation studies with base model(Qwen2.5-VL-7B). Table~\ref{tab:simple_qwen} reports the prompting ablation results, where models are evaluated without rubric guidance to examine the effect of structured rubrics prompting. Tables~\ref{tab:regression}–\ref{tab:multilora+raft} summarize the results of multi-LoRA and RAFT-based ablation experiments, illustrating the contribution of attribute-specific adaptation and feature fusion. Further, Tables~\ref{tab:subset}–\ref{tab:comment+aug} show experiments conducted on the comment-annotated subset, comparing models trained with and without comment signals as well as additional data augmentation through multi-instruction templates. For model training configurations, we fix the image resolution to 224×224 and adopt the following base training configuration: rank = 8, lora\_alpha = 16, learning\_rate = 2e-5, and batch\_size = 16. The corresponding parameter tuning ablations are detailed in Tables~\ref{tab:ab_base}–\ref{tab:ab_r16_la32}. Finally, Table~\ref{tab:aug} reports the comparative results of visual augmentation strategies.

\begin{table}[H]
\centering
\resizebox{0.9\linewidth}{!}{
\small
\begin{tabular}{lcccccc}
\toprule
\textbf{Dimension} & \textbf{SC} & \textbf{PC} & \textbf{ACC} & \textbf{MSE} & \textbf{QWK} \\
\midrule
Realism & 0.640 & 0.680 & 0.780 & 0.220 & 0.621 \\
Deformation & 0.219 & 0.203 & 0.730 & 0.300 & 0.120 \\
Imagination & 0.258 & 0.245 & 0.270 & 0.730 & 0.053 \\
Color Richness & 0.404 & 0.425 & 0.630 & 0.370 & 0.327 \\
Color Contrast & 0.262 & 0.313 & 0.540 & 0.460 & 0.163 \\
Line Combination & 0.168 & 0.169 & 0.550 & 0.480 & 0.145 \\
Line Texture & 0.277 & 0.276 & 0.270 & 1.120 & 0.123 \\
Picture Organization & 0.446 & 0.469 & 0.600 & 0.400 & 0.405 \\
Transformation & 0.283 & 0.276 & 0.670 & 0.330 & 0.217 \\
Average & 0.328 & 0.340 & 0.560 & 0.490 & 0.241 \\
\bottomrule
\end{tabular}
}
\caption{Simple prompting results}
\label{tab:simple_qwen}
\end{table}

\begin{table}[H]
\centering
\resizebox{0.9\linewidth}{!}{
\small
\begin{tabular}{lcccccc}
\toprule
\textbf{Dimension} & \textbf{SC} & \textbf{PC} & \textbf{ACC} & \textbf{MSE} & \textbf{QWK} \\
\midrule
Realism & 0.612 & 0.610 & 0.510 & 0.580 & 0.463 \\
Deformation & 0.337 & 0.393 & 0.330 & 0.670 & 0.223 \\
Imagination & 0.410 & 0.407 & 0.380 & 1.190 & 0.226 \\
Color Richness & 0.535 & 0.561 & 0.360 & 0.790 & 0.348 \\
Color Contrast & 0.523 & 0.539 & 0.280 & 0.930 & 0.298 \\
Line Combination & 0.270 & 0.287 & 0.510 & 0.790 & 0.218 \\
Line Texture & 0.236 & 0.229 & 0.350 & 1.720 & 0.118 \\
Picture Organization & 0.629 & 0.625 & 0.270 & 1.060 & 0.355 \\
Transformation & 0.329 & 0.324 & 0.440 & 0.770 & 0.294 \\
Average & 0.431 & 0.442 & 0.381 & 0.944 & 0.282 \\
\bottomrule
\end{tabular}
}
\caption{Few-shot prompting results (two-shot).}
\label{tab:fewshot_qwen}
\end{table}

\begin{table}[H]
\centering
\resizebox{0.9\linewidth}{!}{
\small
\begin{tabular}{lcccccc}
\toprule
\textbf{Dimension} & \textbf{SC} & \textbf{PC} & \textbf{ACC} & \textbf{MSE} & \textbf{QWK} \\
\midrule
Realism & 0.457 & 0.459 & 0.700 & 0.330 & 0.450 \\
Deformation & 0.524 & 0.518 & 0.770 & 0.230 & 0.511 \\
Imagination & 0.581 & 0.573 & 0.790 & 0.210 & 0.573 \\
Color Richness & 0.615 & 0.617 & 0.660 & 0.370 & 0.615 \\
Color Contrast & 0.517 & 0.519 & 0.690 & 0.340 & 0.519 \\
Line Combination & 0.254 & 0.245 & 0.650 & 0.350 & 0.238 \\
Line Texture & 0.285 & 0.278 & 0.500 & 0.530 & 0.270 \\
Picture Organization & 0.583 & 0.612 & 0.770 & 0.230 & 0.610 \\
Transformation & 0.374 & 0.377 & 0.630 & 0.370 & 0.359 \\
Average & 0.466 & 0.466 & 0.684 & 0.329 & 0.461 \\
\bottomrule
\end{tabular}
}
\caption{Results with LoRA and standard regression}
\label{tab:regression}
\end{table}

\begin{table}[H]
\centering
\resizebox{0.9\linewidth}{!}{
\small
\begin{tabular}{lcccccc}
\toprule
\textbf{Dimension} & \textbf{SC} & \textbf{PC} & \textbf{ACC} & \textbf{MSE} & \textbf{QWK} \\
\midrule
Realism & 0.618 & 0.618 & 0.760 & 0.240 & 0.613 \\
Deformation & 0.553 & 0.514 & 0.810 & 0.220 & 0.505 \\
Imagination & 0.584 & 0.577 & 0.790 & 0.210 & 0.574 \\
Color Richness & 0.662 & 0.661 & 0.670 & 0.330 & 0.654 \\
Color Contrast & 0.573 & 0.600 & 0.710 & 0.290 & 0.600 \\
Line Combination & 0.321 & 0.323 & 0.720 & 0.280 & 0.310 \\
Line Texture & 0.335 & 0.344 & 0.540 & 0.490 & 0.332 \\
Picture Organization & 0.469 & 0.486 & 0.710 & 0.290 & 0.481 \\
Transformation & 0.448 & 0.456 & 0.640 & 0.360 & 0.430 \\
Average & 0.507 & 0.508 & 0.706 & 0.301 & 0.499 \\
\bottomrule
\end{tabular}
}
\caption{Results with multi-LoRA and standard regression}
\label{tab:multilora+regression}
\end{table}

\begin{table}[H]
\centering
\resizebox{0.9\linewidth}{!}{
\small
\begin{tabular}{lcccccc}
\toprule
\textbf{Dimension} & \textbf{SC} & \textbf{PC} & \textbf{ACC} & \textbf{MSE} & \textbf{QWK} \\
\midrule
Realism & 0.545 & 0.541 & 0.750 & 0.250 & 0.537 \\
Deformation & 0.465 & 0.447 & 0.720 & 0.310 & 0.429 \\
Imagination & 0.567 & 0.569 & 0.810 & 0.190 & 0.558 \\
Color Richness & 0.700 & 0.694 & 0.710 & 0.290 & 0.693 \\
Color Contrast & 0.650 & 0.625 & 0.750 & 0.250 & 0.610 \\
Line Combination & 0.367 & 0.356 & 0.760 & 0.240 & 0.330 \\
Line Texture & 0.381 & 0.395 & 0.590 & 0.410 & 0.385 \\
Picture Organization & 0.466 & 0.476 & 0.720 & 0.280 & 0.428 \\
Transformation & 0.523 & 0.519 & 0.720 & 0.280 & 0.497 \\
Average & 0.518 & 0.514 & 0.726 & 0.278 & 0.496 \\
\bottomrule
\end{tabular}
}
\caption{Results with LoRA and RAFT}
\label{tab:raft}
\end{table}

\begin{table}[H]
\centering

\resizebox{0.9\linewidth}{!}{
\small
\begin{tabular}{lcccccc}
\toprule
\textbf{Dimension} & \textbf{SC} & \textbf{PC} & \textbf{ACC} & \textbf{MSE} & \textbf{QWK} \\
\midrule
Realism & 0.762 & 0.800 & 0.890 & 0.106 & 0.804 \\
Deformation & 0.670 & 0.664 & 0.810 & 0.153 & 0.568 \\
Imagination & 0.689 & 0.652 & 0.810 & 0.170 & 0.610 \\
Color Richness & 0.745 & 0.760 & 0.740 & 0.201 & 0.708 \\
Color Contrast & 0.640 & 0.625 & 0.700 & 0.227 & 0.480 \\
Line Combination & 0.517 & 0.512 & 0.730 & 0.199 & 0.396 \\
Line Texture & 0.513 & 0.533 & 0.620 & 0.247 & 0.444 \\
Picture Organization & 0.654 & 0.682 & 0.710 & 0.178 & 0.505 \\
Transformation & 0.643 & 0.650 & 0.720 & 0.237 & 0.574 \\
Average & 0.648 & 0.653 & 0.748 & 0.191 & 0.566 \\
\bottomrule
\end{tabular}
}
\caption{Results with multi-LoRA and RAFT}
\label{tab:multilora+raft}
\end{table}

\begin{table}[H]
\centering
\resizebox{0.9\linewidth}{!}{
\small
\begin{tabular}{lcccccc}
\toprule
\textbf{Dimension} & \textbf{SC} & \textbf{PC} & \textbf{ACC} & \textbf{MSE} & \textbf{QWK} \\
\midrule
Realism & 0.411 & 0.433 & 0.604 & 0.423 & 0.312 \\
Deformation & 0.455 & 0.427 & 0.703 & 0.324 & 0.406 \\
Imagination & 0.202 & 0.198 & 0.441 & 0.559 & 0.120 \\
Color Richness & 0.683 & 0.667 & 0.730 & 0.297 & 0.656 \\
Color Contrast & 0.181 & 0.190 & 0.514 & 0.622 & 0.183 \\
Line Combination & -0.065 & -0.033 & 0.297 & 0.757 & -0.019 \\
Line Texture & 0.172 & 0.164 & 0.369 & 0.847 & 0.108 \\
Picture Organization & 0.386 & 0.402 & 0.441 & 0.586 & 0.301 \\
Transformation & 0.316 & 0.326 & 0.396 & 0.712 & 0.181 \\
Average & 0.304 & 0.308 & 0.499 & 0.570 & 0.250 \\
\bottomrule
\end{tabular}
}
\caption{Results on the subset without comments.}
\label{tab:subset}
\end{table}

\begin{table}[H]
\centering
\resizebox{0.9\linewidth}{!}{
\small
\begin{tabular}{lcccccc}
\toprule
\textbf{Dimension} & \textbf{SC} & \textbf{PC} & \textbf{ACC} & \textbf{MSE} & \textbf{QWK} \\
\midrule
Realism & 0.477 & 0.476 & 0.685 & 0.342 & 0.390 \\
Deformation & 0.322 & 0.334 & 0.459 & 0.541 & 0.227 \\
Imagination & 0.196 & 0.194 & 0.450 & 0.550 & 0.128 \\
Color Richness & 0.603 & 0.622 & 0.523 & 0.505 & 0.508 \\
Color Contrast & 0.502 & 0.564 & 0.360 & 0.775 & 0.355 \\
Line Combination & 0.067 & 0.078 & 0.631 & 0.450 & 0.051 \\
Line Texture & 0.219 & 0.193 & 0.468 & 0.802 & 0.147 \\
Picture Organization & 0.451 & 0.472 & 0.613 & 0.387 & 0.349 \\
Transformation & 0.359 & 0.340 & 0.450 & 0.739 & 0.245 \\
Average & 0.355 & 0.364 & 0.516 & 0.566 & 0.267 \\
\bottomrule
\end{tabular}
}
\caption{Results on the subset with comments.}
\label{tab:comment}
\end{table}

\vspace{-1.2 cm}

\begin{table}[H]
\centering
\resizebox{0.9\linewidth}{!}{
\small
\begin{tabular}{lcccccc}
\toprule
\textbf{Dimension} & \textbf{SC} & \textbf{PC} & \textbf{ACC} & \textbf{MSE} & \textbf{QWK} \\
\midrule
Realism & 0.654 & 0.640 & 0.739 & 0.212 & 0.485 \\
Deformation & 0.494 & 0.495 & 0.396 & 0.481 & 0.158 \\
Imagination & 0.531 & 0.359 & 0.450 & 0.496 & 0.132 \\
Color Richness & 0.704 & 0.720 & 0.396 & 0.599 & 0.424 \\
Color Contrast & 0.615 & 0.661 & 0.640 & 0.297 & 0.545 \\
Line Combination & 0.081 & 0.120 & 0.486 & 0.729 & 0.040 \\
Line Texture & 0.259 & 0.250 & 0.505 & 0.465 & 0.137 \\
Picture Organization & 0.541 & 0.589 & 0.622 & 0.326 & 0.160 \\
Transformation & 0.420 & 0.411 & 0.486 & 0.617 & 0.252 \\
Average & 0.478 & 0.472 & 0.525 & 0.469 & 0.259 \\
\bottomrule
\end{tabular}
}
\caption{Results on the subset with comments and augment.}
\label{tab:comment+aug}
\end{table}

\vspace{-1.2 cm}

\begin{table}[H]
\centering
\resizebox{0.9\linewidth}{!}{
\small
\begin{tabular}{lcccccc}
\toprule
\textbf{Dimension} & \textbf{SC} & \textbf{PC} & \textbf{ACC} & \textbf{MSE} & \textbf{QWK} \\
\midrule
Base & 0.545 & 0.546 & 0.728 & 0.275 & 0.530 \\
Rotation & 0.385 & 0.385 & 0.727 & 0.280 & 0.353 \\
Hflip & 0.458 & 0.475 & 0.731 & 0.276 & 0.466 \\
Blur & 0.523 & 0.520 & 0.733 & 0.270 & 0.508 \\
Colorjitter & 0.582 & 0.585 & 0.743 & 0.263 & 0.558 \\
\bottomrule
\end{tabular}
}
\caption{Results under augmentation strategies.}
\label{tab:aug}
\end{table}

\vspace{-1.2 cm}

\begin{table}[H]
\centering
\resizebox{1\linewidth}{!}{
\small
\begin{tabular}{lcccccc}
\toprule
\textbf{Dimension} & \textbf{PC (w/o C)} & \textbf{RMSE (w/o C)} & \textbf{QWK (w/o C)} &
\textbf{PC (w C)} & \textbf{RMSE (w C)} & \textbf{QWK (w C)} \\
\midrule
Realism               & 0.726 & 0.387 & 0.722 & 0.697 & 0.387 & 0.519 \\
Deformation           & 0.570 & 0.469 & 0.557 & 0.599 & 0.434 & 0.512 \\
Imagination           & 0.575 & 0.469 & 0.559 & 0.677 & 0.374 & 0.664 \\
Color Richness       & 0.679 & 0.538 & 0.679 & 0.759 & 0.450 & 0.722 \\
Color Contrast       & 0.591 & 0.529 & 0.586 & 0.660 & 0.468 & 0.683 \\
Line Combination     & 0.325 & 0.490 & 0.290 & 0.445 & 0.455 & 0.332 \\
Line Texture         & 0.486 & 0.600 & 0.468 & 0.442 & 0.555 & 0.380 \\
Picture Organization & 0.640 & 0.447 & 0.627 & 0.641 & 0.435 & 0.506 \\
Transformation        & 0.603 & 0.500 & 0.599 & 0.587 & 0.481 & 0.537 \\
Average               & 0.577 & 0.496 & 0.565 & 0.612 & 0.452 & 0.539 \\
\bottomrule
\end{tabular}
}
\caption{Results on the full dataset w/wo comments.}
\label{tab:comment_full}
\end{table}

\vspace{-1.2 cm}

\begin{table}[H]
\centering
\resizebox{0.9\linewidth}{!}{
\small
\begin{tabular}{lcccccc}
\toprule
\textbf{Dimension} & \textbf{SC} & \textbf{PC} & \textbf{ACC} & \textbf{MSE} & \textbf{QWK} \\
\midrule
Realism & 0.604 & 0.599 & 0.780 & 0.220 & 0.593 \\
Deformation & 0.526 & 0.488 & 0.780 & 0.250 & 0.480 \\
Imagination & 0.618 & 0.611 & 0.810 & 0.190 & 0.604 \\
Color Richness & 0.688 & 0.700 & 0.670 & 0.330 & 0.690 \\
Color Contrast & 0.716 & 0.722 & 0.790 & 0.210 & 0.710 \\
Line Combination & 0.341 & 0.342 & 0.730 & 0.270 & 0.324 \\
Line Texture & 0.437 & 0.455 & 0.570 & 0.430 & 0.392 \\
Picture Organization & 0.414 & 0.434 & 0.690 & 0.310 & 0.426 \\
Transformation & 0.565 & 0.568 & 0.730 & 0.270 & 0.557 \\
Average & 0.546 & 0.546 & 0.728 & 0.276 & 0.531 \\
\bottomrule
\end{tabular}
}
\caption{Performance under base configuration.}
\label{tab:ab_base}
\end{table}

\vspace{-1.2 cm}

\begin{table}[H]
\centering
\resizebox{0.9\linewidth}{!}{
\small
\begin{tabular}{lcccccc}
\toprule
\textbf{Dimension} & \textbf{SC} & \textbf{PC} & \textbf{ACC} & \textbf{MSE} & \textbf{QWK} \\
\midrule
Realism & 0.602 & 0.609 & 0.770 & 0.230 & 0.603 \\
Deformation & 0.482 & 0.447 & 0.770 & 0.260 & 0.442 \\
Imagination & 0.652 & 0.644 & 0.830 & 0.170 & 0.641 \\
Color Richness & 0.665 & 0.681 & 0.700 & 0.300 & 0.680 \\
Color Contrast & 0.667 & 0.672 & 0.760 & 0.240 & 0.671 \\
Line Combination & 0.320 & 0.310 & 0.730 & 0.270 & 0.298 \\
Line Texture & 0.408 & 0.412 & 0.520 & 0.480 & 0.325 \\
Picture Organization & 0.485 & 0.490 & 0.720 & 0.280 & 0.461 \\
Transformation & 0.507 & 0.507 & 0.720 & 0.280 & 0.496 \\
Average & 0.532 & 0.530 & 0.724 & 0.279 & 0.513 \\
\bottomrule
\end{tabular}
}
\caption{Performance under base training configuration with batch\_size = 4.}
\label{tab:ab_bs4}
\end{table}

\vspace{-0.4cm}

\begin{table}[H]
\centering
\resizebox{0.9\linewidth}{!}{
\small
\begin{tabular}{lcccccc}
\toprule
\textbf{Dimension} & \textbf{SC} & \textbf{PC} & \textbf{ACC} & \textbf{MSE} & \textbf{QWK} \\
\midrule
Realism & 0.563 & 0.558 & 0.760 & 0.240 & 0.556 \\
Deformation & 0.451 & 0.418 & 0.740 & 0.290 & 0.410 \\
Imagination & 0.634 & 0.627 & 0.820 & 0.180 & 0.622 \\
Color Richness & 0.695 & 0.710 & 0.730 & 0.270 & 0.706 \\
Color Contrast & 0.613 & 0.637 & 0.740 & 0.260 & 0.636 \\
Line Combination & 0.340 & 0.329 & 0.740 & 0.260 & 0.313 \\
Line Texture & 0.391 & 0.385 & 0.560 & 0.440 & 0.363 \\
Picture Organization & 0.542 & 0.552 & 0.750 & 0.250 & 0.545 \\
Transformation & 0.554 & 0.557 & 0.730 & 0.270 & 0.537 \\
Average & 0.532 & 0.530 & 0.730 & 0.273 & 0.521 \\
\bottomrule
\end{tabular}
}
\caption{Performance under base training configuration with batch\_size = 24.}
\label{tab:ab_bs24}
\end{table}

\vspace{-0.4cm}

\begin{table}[H]
\centering
\resizebox{0.9\linewidth}{!}{
\small
\begin{tabular}{lcccccc}
\toprule
\textbf{Dimension} & \textbf{SC} & \textbf{PC} & \textbf{ACC} & \textbf{MSE} & \textbf{QWK} \\
\midrule
Realism & 0.478 & 0.481 & 0.710 & 0.290 & 0.463 \\
Deformation & 0.474 & 0.440 & 0.780 & 0.250 & 0.431 \\
Imagination & 0.629 & 0.621 & 0.810 & 0.190 & 0.610 \\
Color Richness & 0.726 & 0.737 & 0.740 & 0.260 & 0.734 \\
Color Contrast & 0.691 & 0.697 & 0.780 & 0.220 & 0.693 \\
Line Combination & 0.338 & 0.326 & 0.730 & 0.270 & 0.311 \\
Line Texture & 0.407 & 0.402 & 0.560 & 0.440 & 0.363 \\
Picture Organization & 0.510 & 0.519 & 0.740 & 0.260 & 0.497 \\
Transformation & 0.563 & 0.569 & 0.750 & 0.250 & 0.556 \\
Average & 0.535 & 0.532 & 0.733 & 0.270 & 0.518 \\
\bottomrule
\end{tabular}
}
\caption{Performance under base training configuration with learning\_rate = 1e-5.}
\label{tab:ab_lr1e-5}
\end{table}

\vspace{-0.4cm}

\begin{table}[H]
\centering
\resizebox{0.9\linewidth}{!}{
\small
\begin{tabular}{lcccccc}
\toprule
\textbf{Dimension} & \textbf{SC} & \textbf{PC} & \textbf{ACC} & \textbf{MSE} & \textbf{QWK} \\
\midrule
Realism & 0.642 & 0.633 & 0.800 & 0.200 & 0.630 \\
Deformation & 0.455 & 0.422 & 0.770 & 0.260 & 0.415 \\
Imagination & 0.654 & 0.649 & 0.840 & 0.160 & 0.649 \\
Color Richness & 0.713 & 0.727 & 0.740 & 0.260 & 0.719 \\
Color Contrast & 0.664 & 0.668 & 0.760 & 0.270 & 0.662 \\
Line Combination & 0.230 & 0.238 & 0.720 & 0.280 & 0.197 \\
Line Texture & 0.385 & 0.420 & 0.530 & 0.470 & 0.336 \\
Picture Organization & 0.494 & 0.498 & 0.710 & 0.290 & 0.492 \\
Transformation & 0.442 & 0.444 & 0.680 & 0.320 & 0.425 \\
Average & 0.520 & 0.522 & 0.728 & 0.279 & 0.503 \\
\bottomrule
\end{tabular}
}
\caption{Performance under base training configuration with learning\_rate = 5e-5.}
\label{tab:ab_lr5e-5}
\end{table}

\vspace{-0.4cm}

\begin{table}[H]
\centering
\resizebox{0.9\linewidth}{!}{
\small
\begin{tabular}{lcccccc}
\toprule
\textbf{Dimension} & \textbf{SC} & \textbf{PC} & \textbf{ACC} & \textbf{MSE} & \textbf{QWK} \\
\midrule
Realism & 0.583 & 0.576 & 0.770 & 0.230 & 0.574 \\
Deformation & 0.331 & 0.308 & 0.750 & 0.280 & 0.268 \\
Imagination & 0.616 & 0.612 & 0.820 & 0.180 & 0.611 \\
Color Richness & 0.724 & 0.736 & 0.750 & 0.250 & 0.735 \\
Color Contrast & 0.645 & 0.620 & 0.750 & 0.250 & 0.608 \\
Line Combination & 0.240 & 0.237 & 0.730 & 0.270 & 0.216 \\
Line Texture & 0.296 & 0.286 & 0.550 & 0.480 & 0.283 \\
Picture Organization & 0.535 & 0.543 & 0.750 & 0.250 & 0.529 \\
Transformation & 0.567 & 0.568 & 0.740 & 0.260 & 0.549 \\
Average & 0.504 & 0.498 & 0.734 & 0.272 & 0.486 \\
\bottomrule
\end{tabular}
}
\caption{Performance under base training configuration with rank = 4 and lora\_alpha = 16.}
\label{tab:ab_r4_la8}
\end{table}

\vspace{-0.4cm}

\begin{table}[H]
\centering
\resizebox{0.9\linewidth}{!}{
\small
\begin{tabular}{lcccccc}
\toprule
\textbf{Dimension} & \textbf{SC} & \textbf{PC} & \textbf{ACC} & \textbf{MSE} & \textbf{QWK} \\
\midrule
Realism & 0.528 & 0.527 & 0.740 & 0.260 & 0.519 \\
Deformation & 0.496 & 0.460 & 0.770 & 0.260 & 0.454 \\
Imagination & 0.649 & 0.645 & 0.840 & 0.160 & 0.643 \\
Color Richness & 0.732 & 0.743 & 0.750 & 0.250 & 0.741 \\
Color Contrast & 0.613 & 0.627 & 0.720 & 0.280 & 0.627 \\
Line Combination & 0.263 & 0.258 & 0.740 & 0.260 & 0.231 \\
Line Texture & 0.341 & 0.344 & 0.550 & 0.450 & 0.325 \\
Picture Organization & 0.524 & 0.523 & 0.740 & 0.260 & 0.513 \\
Transformation & 0.578 & 0.579 & 0.740 & 0.260 & 0.569 \\
Average & 0.525 & 0.523 & 0.732 & 0.271 & 0.513 \\
\bottomrule
\end{tabular}
}
\caption{Performance under base training configuration with rank = 16 and lora\_alpha = 32.}
\label{tab:ab_r16_la32}
\end{table}

\vspace{-0.4cm}

\begin{table*}[t]
\centering
\resizebox{1.0\linewidth}{!}{
\small
\begin{tabular}{lccccccccc}
\toprule
\textbf{Dimension} &
\textbf{PC\_gt\_T1} & \textbf{PC\_gt\_T2} & \textbf{PC\_T1\_T2} &
\textbf{ICC\_gt\_T1} & \textbf{ICC\_gt\_T2} & \textbf{ICC\_T1\_T2} &
\textbf{$\alpha$\_gt\_T1} & \textbf{$\alpha$\_gt\_T2} & \textbf{$\alpha$\_T1\_T2} \\
\midrule
Realism              & 0.7611 & 0.7433 & 0.5602 & 0.7383 & 0.7335 & 0.5581 & 0.7906 & 0.7592 & 0.6050 \\
Deformation          & 0.4806 & 0.5521 & 0.2744 & 0.3935 & 0.4424 & 0.2744 & 0.4072 & 0.4484 & 0.2377 \\
Imagination          & 0.5013 & 0.4874 & 0.2225 & 0.4269 & 0.4234 & 0.2224 & 0.4604 & 0.4482 & 0.2145 \\
Color Richness      & 0.8655 & 0.8701 & 0.7606 & 0.8612 & 0.8647 & 0.7607 & 0.8567 & 0.8613 & 0.7509 \\
Color Contrast      & 0.8357 & 0.8313 & 0.6971 & 0.8254 & 0.8199 & 0.6968 & 0.8274 & 0.8345 & 0.7091 \\
Line Combination    & 0.5449 & 0.5584 & 0.2841 & 0.4737 & 0.4707 & 0.2837 & 0.4967 & 0.5122 & 0.2890 \\
Line Texture        & 0.6217 & 0.6246 & 0.3612 & 0.5728 & 0.5652 & 0.3610 & 0.6047 & 0.5916 & 0.3554 \\
Picture Organization & 0.7425 & 0.7145 & 0.5191 & 0.7102 & 0.6831 & 0.5191 & 0.7614 & 0.7344 & 0.5672 \\
Transformation       & 0.5311 & 0.5190 & 0.2666 & 0.4657 & 0.4589 & 0.2667 & 0.4928 & 0.4872 & 0.2530 \\
\bottomrule
\end{tabular}
}
\caption{Inter-rater agreement statistics across dimensions, including Pearson correlation (PC), intraclass correlation coefficient (ICC), and Krippendorff’s alpha ($\alpha$).}
\label{tab:agreement}
\end{table*}

\subsection{Annotation Details}

For annotation process, evaluating children’s artwork inherently involves conceptual overlap among dimensions and a degree of subjectivity, as many aesthetic and creative attributes are conceptually adjacent rather than strictly orthogonal. Our rubric is therefore designed to provide a practically usable and pedagogically meaningful decomposition of children’s artistic expression, balancing fine-grained descriptive power with the need for consistent large-scale annotation.

To ensure reliability, we employed a multi-stage annotation protocol in which initial ratings were iteratively refined and verified, and final scores were consolidated by expert annotators. This process aimed to standardize rubric interpretation and reduce variability across annotators. The resulting framework reflects a deliberate design choice: dimensions are more specific than broad holistic categories yet not as atomized as highly detailed taxonomies, enabling nuanced assessment without sacrificing feasibility. Some degree of inter-dimension correlation is structurally expected given the conceptual nature of the attributes. In art-education practice, for instance, Imagination, Deformation, and Transformation often co-occur in children’s creative processes, while remaining qualitatively distinct (e.g., imagination foregrounding novelty, transformation emphasizing systematic modification). Furthermore, analyses of model behavior and dataset structure (Section ~\ref{Qual Analysis}) support these distinctions.

To quantify annotation reliability, we computed inter-rater agreement prior to expert consolidation, as the results shown in Table ~\ref{tab:agreement}. Concrete perceptual attributes such as color richness and color contrast show high agreement ($\alpha$ = 0.83–0.86), whereas more abstract, intent-based attributes—including imagination, deformation, and transformation—exhibit lower agreement. This pattern reflects the inherent interpretive difficulty of these constructs. Importantly, the labels used in our experiments derive from the multi-stage consolidation workflow, which explicitly resolves disagreements and improves consistency relative to the initial independent ratings.

\subsection{Qualitative results}
\label{app:qua}

\begin{figure*}[!ht]
  \centering
  \includegraphics[scale=0.4]{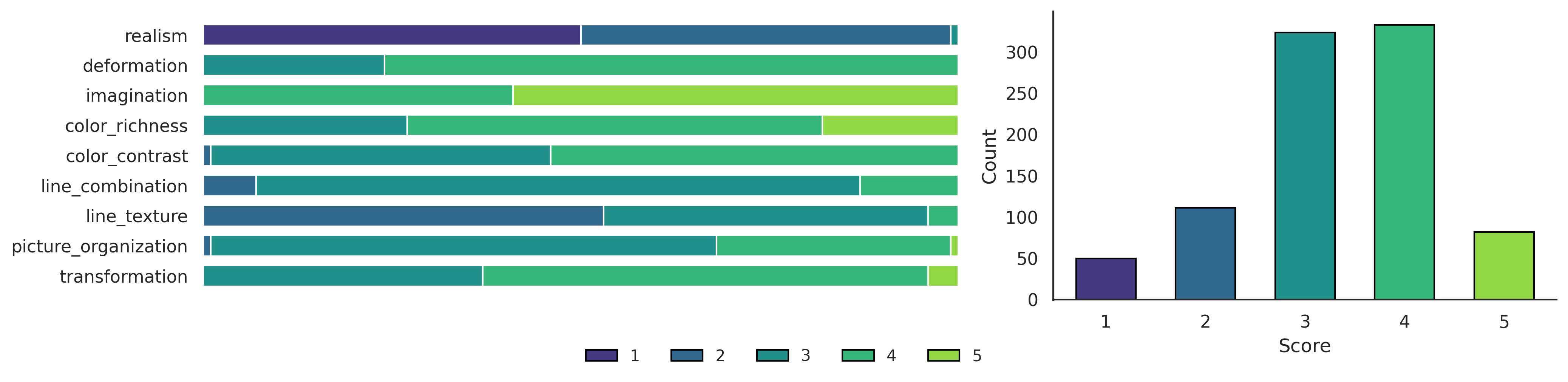}
  \caption{The distribution for our base model predictions with each dimension (Left) and overall results (Right).}
  \label{fig:fig6}
\end{figure*}

Figure~\ref{fig:fig6} illustrates the distribution of our base model’s predictions across each artistic dimension and the overall distribution. When compared with the ground-truth score distributions shown in Figure~\ref{fig:distribution}, the predicted results exhibit a broadly consistent statistical pattern, indicating that the model captures the relative score tendencies rather than producing random or biased estimations. This similarity suggests that the model has learned to approximate human scoring behavior in a stable and interpretable manner.

\begin{figure}[H]
  \centering
  \includegraphics[scale=0.25]{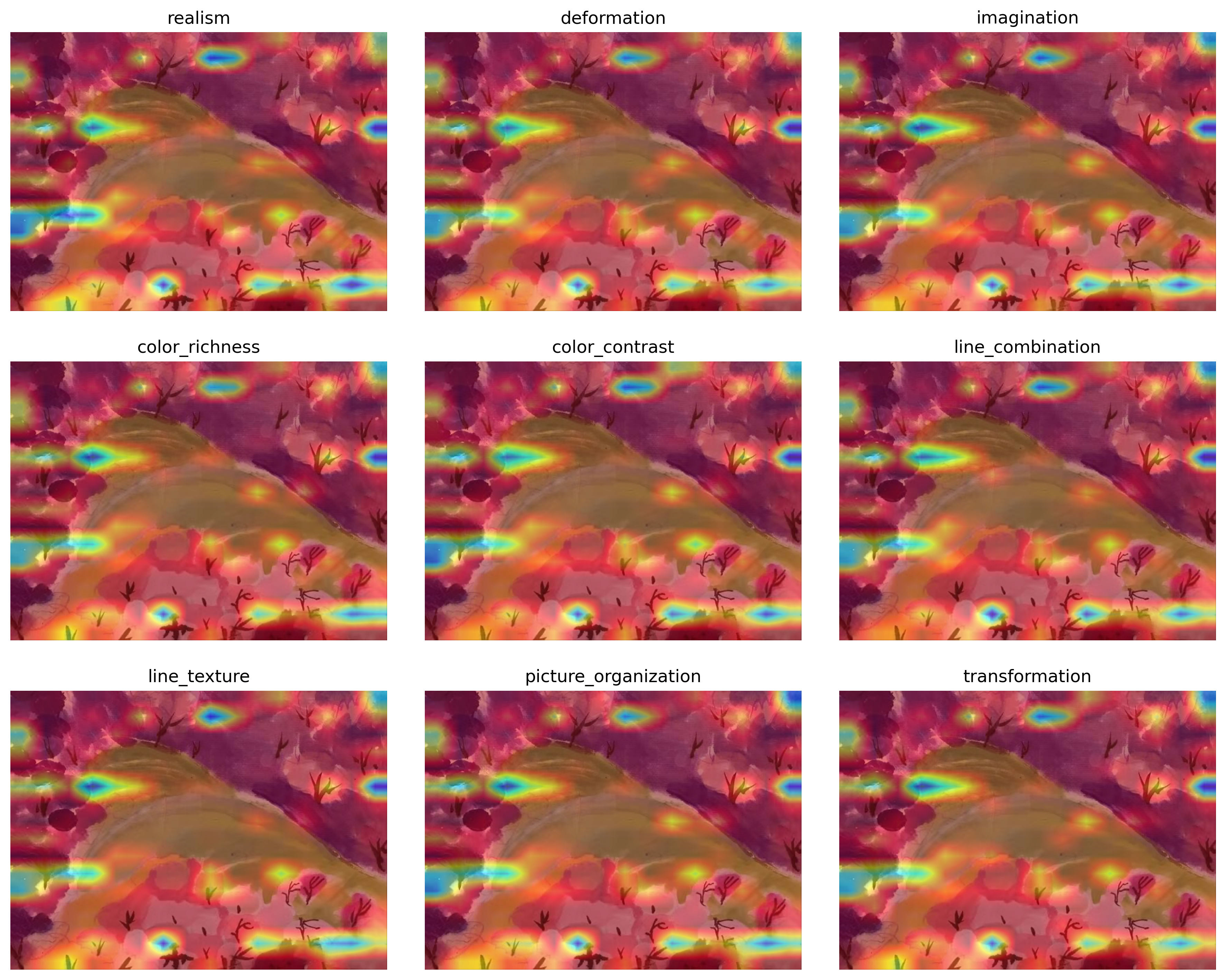}
  \caption{Spatial visual attention visualized by Gradient-CAM for an example image.}
  \label{fig:fig7}
\end{figure}

To further explore the model’s perceptual reasoning, Figure~\ref{fig:fig7} presents spatial visual attention maps generated via Gradient-CAM. We observe that models for different dimensions attend to distinct regions and details, while maintaining some shared focus on general visual structures. This indicates both specialization and commonality in attention mechanisms across aesthetic dimensions.

\begin{figure}[H]
  \centering
  \includegraphics[scale=0.3]{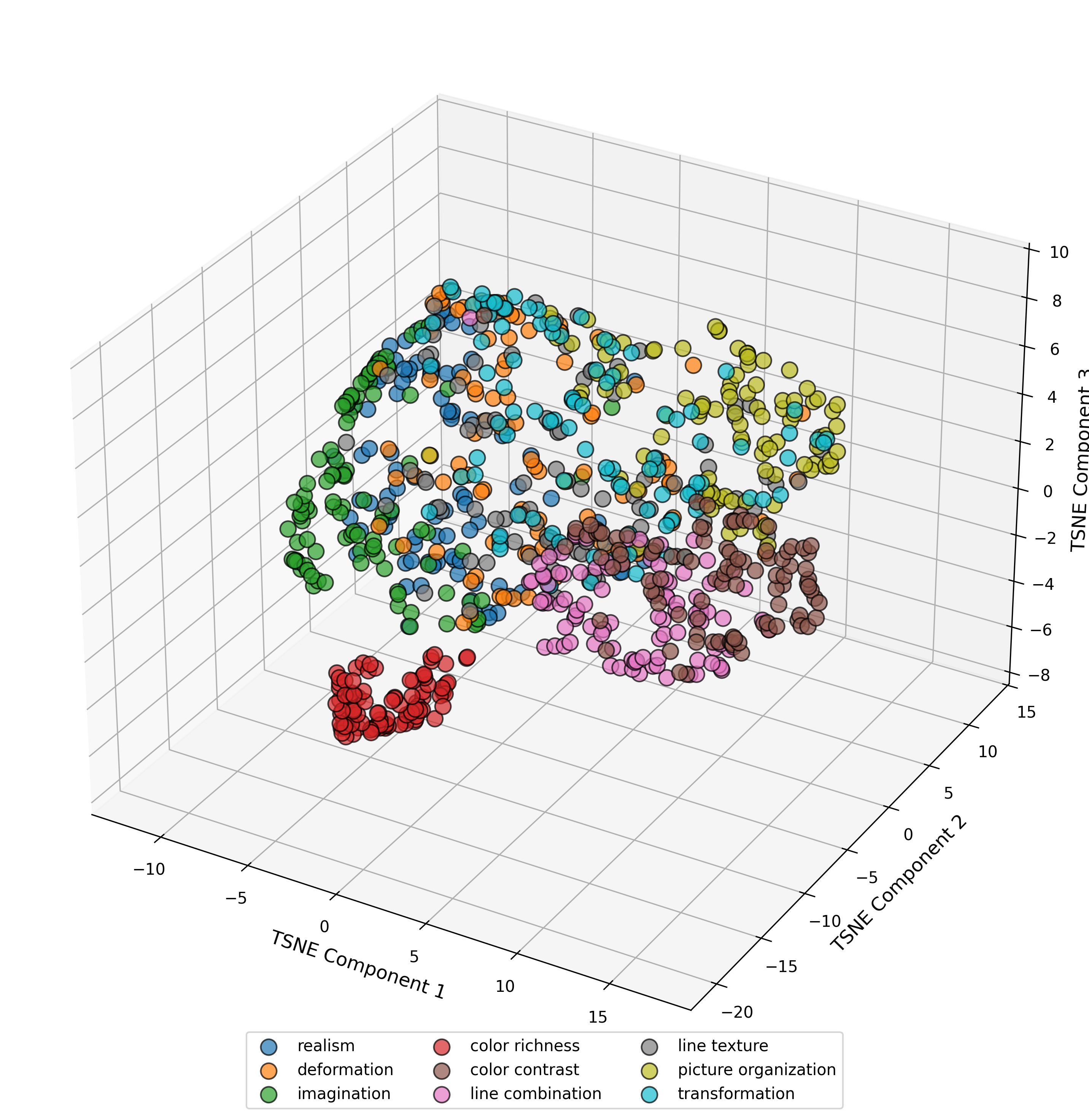}
  \caption{t-SNE projection of final-layer attention features across dimensions..
}
  \label{fig:fig8}
\end{figure}

% To investigate the representational structure underlying these attention patterns, we conduct an attention-based clustering analysis (Figure~\ref{fig:fig8}). We process the set of 100 test images, from which statistical attention features were extracted from the final layer and projected into a shared three-dimensional embedding space using t-SNE. The clusters corresponding to different dimensions (e.g., Color Richness, Imagination, Picture Organization) exhibit clear separability, suggesting that the models capture distinct perceptual representations aligned with artistic attributes. However, the Line Texture cluster (in gray) appears notably dispersed, indicating weaker intra-dimensional consistency—an observation that aligns with its relatively lower quantitative performance (see Figure~\ref{fig:fig2} and Table~\ref{tab:qwen_all}). This suggests that Line Texture remains a challenging aspect for the model to comprehend, likely due to its subtle and fine-grained visual characteristics.

\begin{figure}[H]
  \centering
  \includegraphics[scale=0.25]{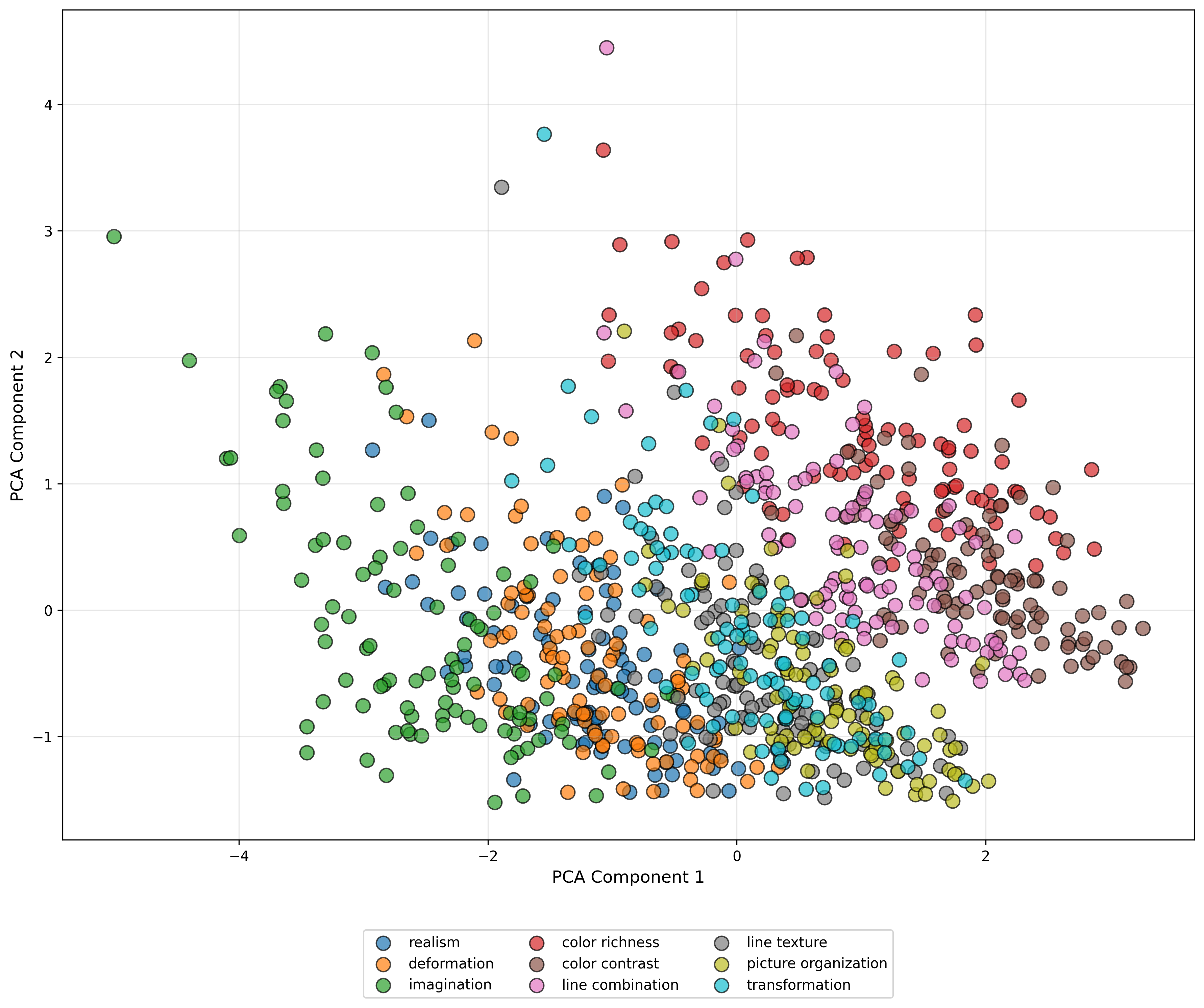}
  \caption{PCA projection of final-layer attention features across dimensions.}
  \label{fig:fig9}
\end{figure}

To investigate the representational structure underlying these attention patterns, we conduct an attention-based clustering analysis (Figure~\ref{fig:fig8}-~\ref{fig:fig9}). We process the set of 100 test images, from which statistical attention features were extracted from the final layer and projected into a shared embedding space using t-SNE and PCA projection. The clusters corresponding to different dimensions (e.g., Color Richness, Imagination, Picture Organization) exhibit clear separability, suggesting that the models capture distinct perceptual representations aligned with artistic attributes. However, the Line Texture cluster (in gray) appears notably dispersed, indicating weaker intra-dimensional consistency—an observation that aligns with its relatively lower quantitative performance (see Figure~\ref{fig:fig2} and Table~\ref{tab:qwen_all}). This suggests that Line Texture remains a challenging aspect for the model to comprehend, likely due to its subtle and fine-grained visual characteristics. 

Overall, these results indicate that the learned feature representations corresponding to different dimensions are not overly abstract or heavily overlapping. Across data-level patterns, model performance trends, and feature-space structures, the results collectively support the necessity of explicitly modeling these dimensions. In the context of our task, this dimensional formulation enables more faithful modeling of the diverse perceptual and cognitive aspects involved in children’s artwork assessment, thereby supporting more interpretable and fine-grained evaluation outcomes.

\end{document}